%% file: main.tex
\documentclass{article}
\pdfoutput=1

\usepackage[final, nonatbib]{neurips_2024}
\usepackage[utf8]{inputenc} % allow utf-8 input
\usepackage[T1]{fontenc}    % use 8-bit T1 fonts
\usepackage{hyperref}       % hyperlinks
\usepackage{url}            % simple URL typesetting
\usepackage{booktabs}       % professional-quality tables
\usepackage{amsfonts}       % blackboard math symbols
\usepackage{nicefrac}       % compact symbols for 1/2, etc.
\usepackage{microtype}      % microtypography
\usepackage{xcolor}         % colors

\usepackage[english]{babel}

\usepackage{amsmath}
\usepackage{graphicx}
\usepackage{multirow}
\usepackage{colortbl}

\usepackage{algorithm}
\usepackage{algpseudocode}

\usepackage{subcaption}

\definecolor{highlight}{rgb}{0.92,0.97,1} % light blue

\title{
Cherry on Top: Parameter Heterogeneity and Quantization in Large Language Models
}

\author{
    Wanyun Cui\textsuperscript{*$\dagger$ $\ddagger$}, \,  
    Qianle Wang\thanks{Equal contribution}\;\,\textsuperscript{$\dagger$} \\
    \textsuperscript{$\dagger$}Shanghai University of Finance and Economics \\
    \textsuperscript{$\ddagger$}MoE Key Laboratory of Interdisciplinary Research of Computation and Economics, \\ Shanghai University of Finance and Economics \\
    \texttt{cui.wanyun@sufe.edu.cn}, \texttt{wql20000111@stu.sufe.edu.cn} 
}

\begin{document}

\maketitle

\begin{abstract}
  % The abstract paragraph should be indented \nicefrac{1}{2}~inch (3~picas) on
  % both the left- and right-hand margins. Use 10~point type, with a vertical
  % spacing (leading) of 11~points.  The word \textbf{Abstract} must be centered,
  % bold, and in point size 12. Two line spaces precede the abstract. The abstract
  % must be limited to one paragraph.
  This paper reveals the phenomenon of parameter heterogeneity in large language models (LLMs). We find that a small subset of ``cherry'' parameters exhibit a disproportionately large influence on model performance, while the vast majority of parameters have minimal impact. This heterogeneity is found to be prevalent across different model families, scales, and types. Motivated by this observation, we propose CherryQ, a novel quantization method that unifies the optimization of mixed-precision parameters. CherryQ identifies and preserves the critical cherry parameters in high precision while aggressively quantizing the remaining parameters to low precision. Extensive experiments demonstrate the effectiveness of CherryQ. CherryQ outperforms existing quantization approaches in terms of perplexity and downstream task performance. Notably, our 3-bit quantized Vicuna-1.5 exhibits competitive performance compared to their 16-bit counterparts. %Our 2-bit quantization method significantly outperforms the SOTA approaches. %These findings highlight the potential of CherryQ for enabling efficient deployment of LLMs by taking advantage of parameter heterogeneity.
\end{abstract}

\input{intro}
\input{related}

\input{measure}
\input{method}
\input{phenomenon}

\input{experiments}

\input{conclu}

\bibliographystyle{plain}
\bibliography{heterogeneity}

%%%%%%%%%%%%%%%%%%%%%%%%%%%%%%%%%%%%%%%%%%%%%%%%%%%%%%%%%%%%

% \appendix

% \section{Appendix / supplemental material}

% Optionally include supplemental material (complete proofs, additional experiments and plots) in appendix.
% All such materials \textbf{SHOULD be included in the main submission.}

\newpage
\input{appendix}

\input{checklist}

\end{document}

%% file: intro.tex
\section{Introduction}

The rapid development of large language models (LLMs) has increased the demand of efficient deployment in various environments~\cite{achiam2023gpt,bai2023qwen,jiang2023mistral,touvron2023llama}. However, the parameter size poses significant challenges for GPU memory requirements. Quantization, which reduces the bit-width of model parameters, has emerged as a solution to alleviate memory constraints of LLM deployment~\cite{kim2023squeezellm,krishnamoorthi2018quantizing,liu2023llm,wei2022outlier,xiao2023smoothquant}.

\begin{figure}[htbp]
    \centering
    \begin{subfigure}[b]{0.32\textwidth}
        \centering
        \includegraphics[width=\textwidth]{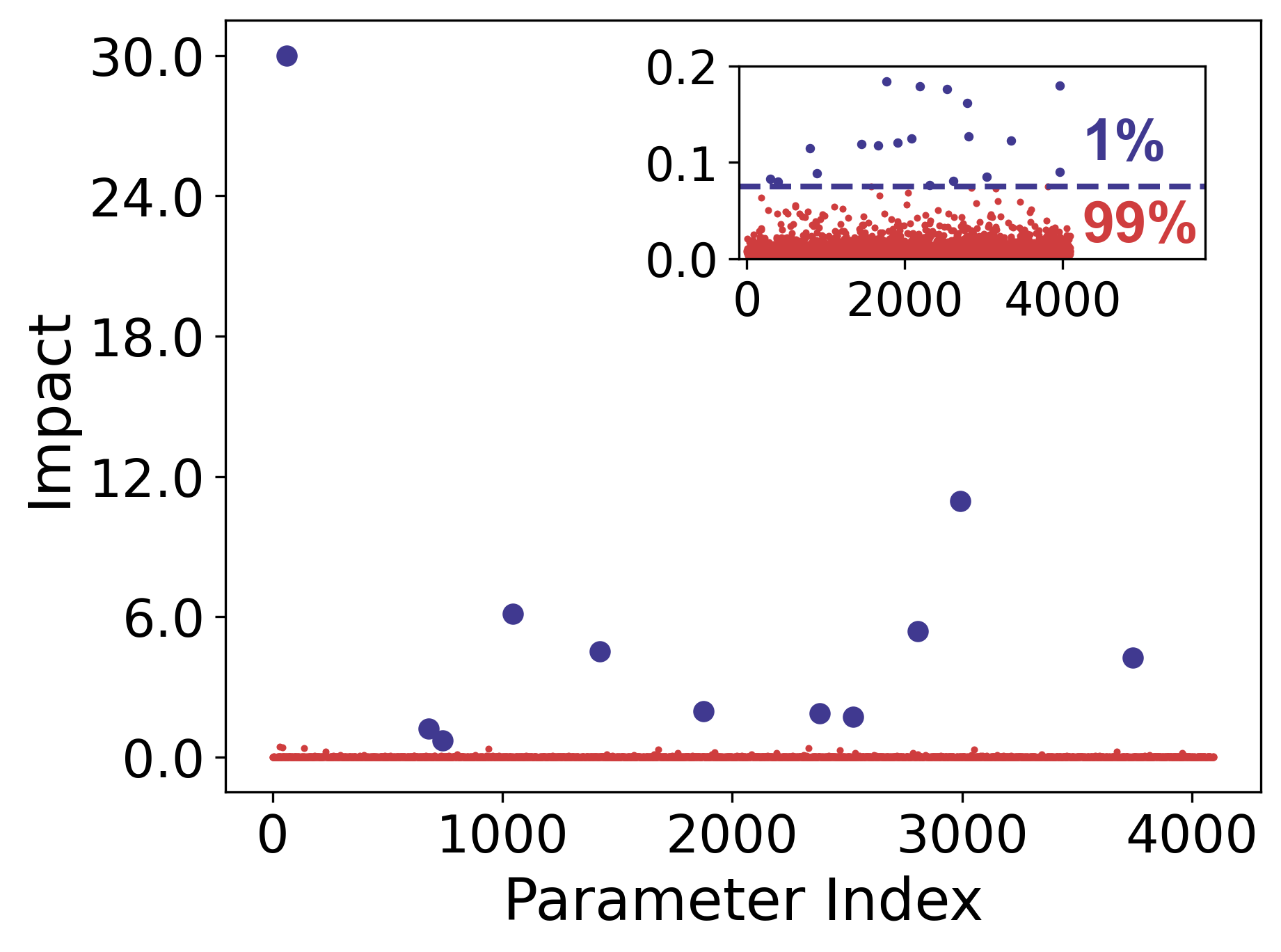}
        \caption{LLaMA2 7B \\ {\tt layers.15.self\_attn.v\_proj}}
        \label{fig:param_heterogeneity:llama7b}
    \end{subfigure}
    \hfill
    \begin{subfigure}[b]{0.315\textwidth}
        \centering
        \includegraphics[width=\textwidth]{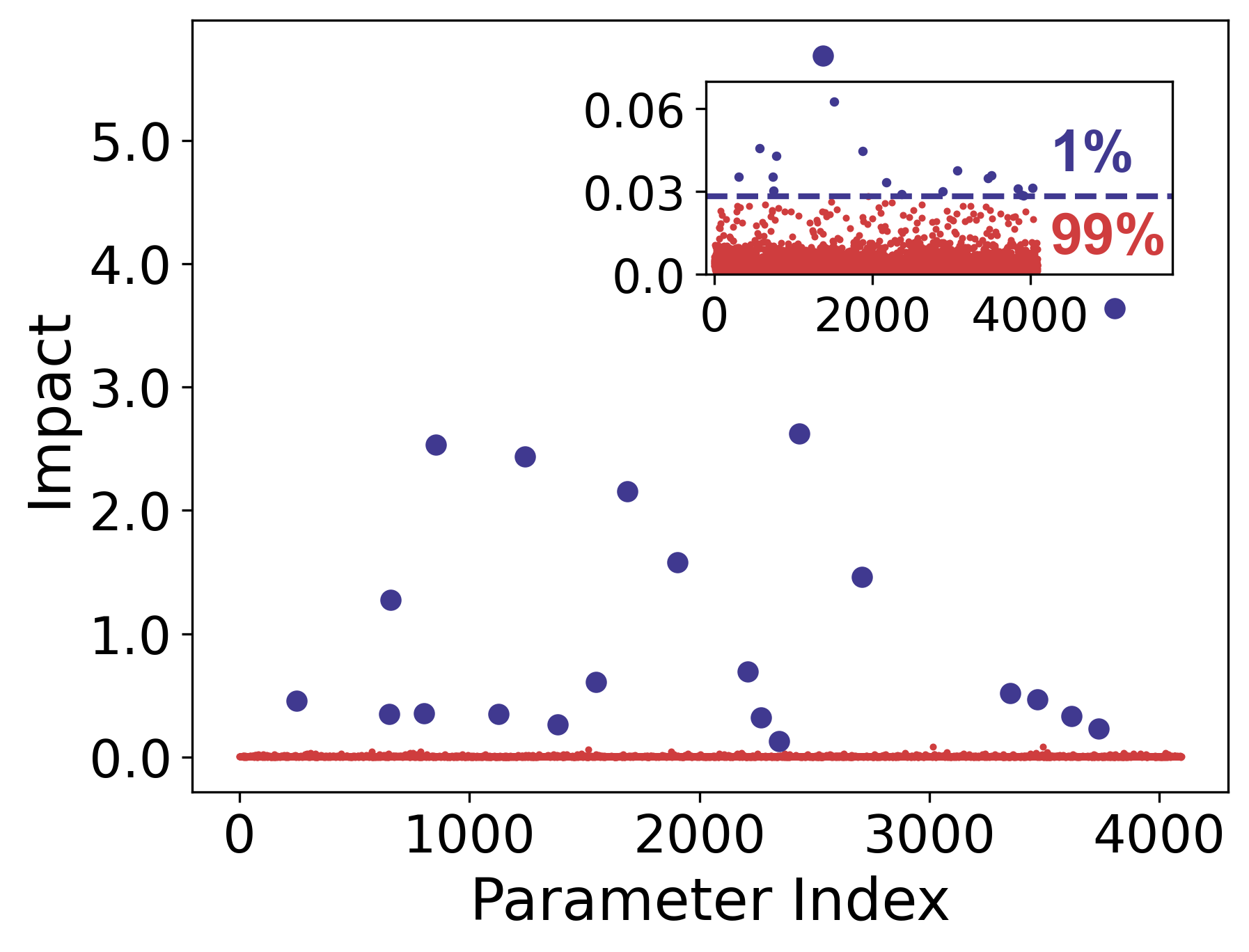}
        \caption{LLaMA2 13B \\ {\tt layers.15.self\_attn.v\_proj}}
        \label{fig:param_heterogeneity:llama13b}
    \end{subfigure}
    \hfill
    \begin{subfigure}[b]{0.335\textwidth}
        \centering
        \includegraphics[width=\textwidth]{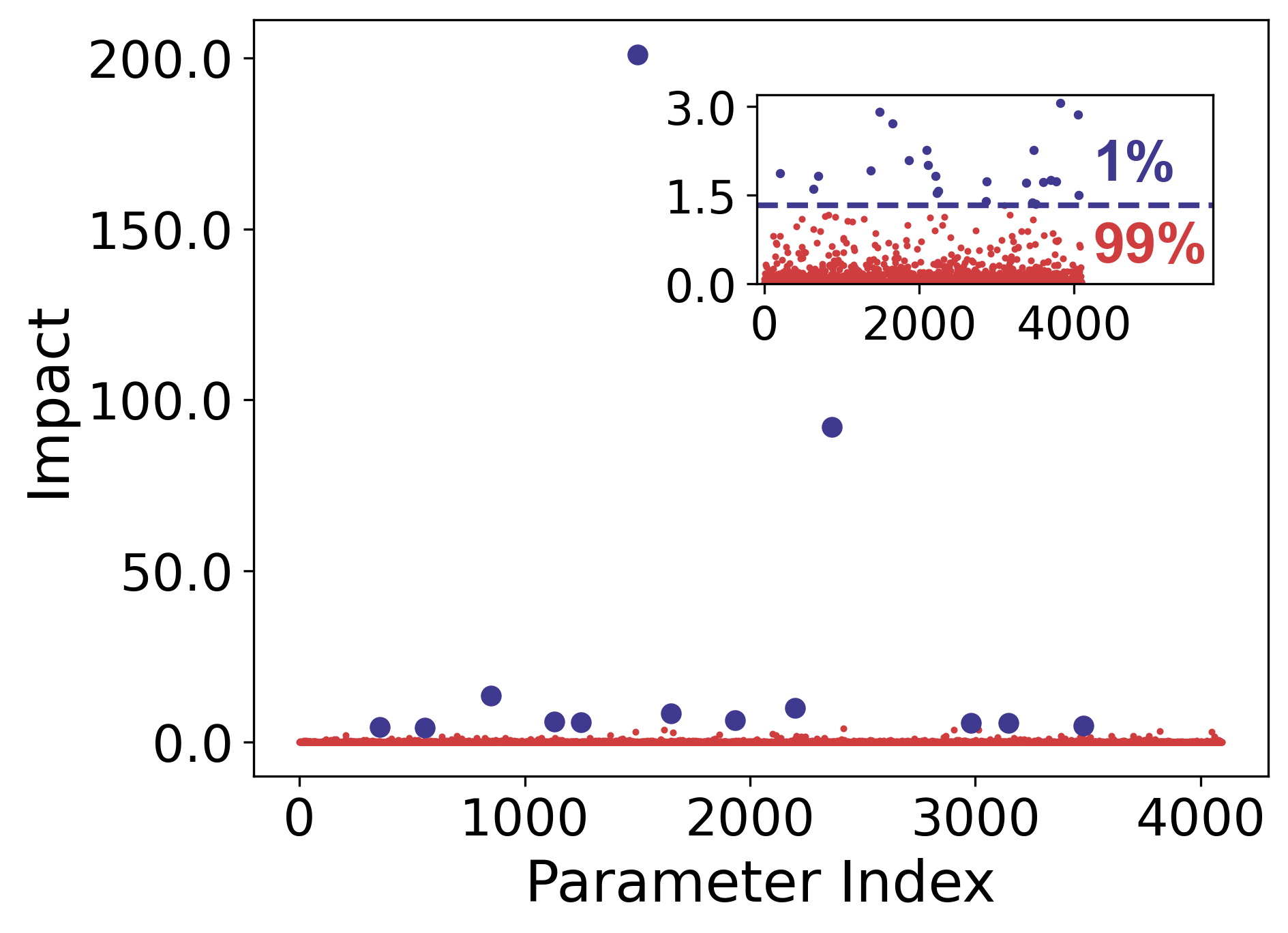}
        \caption{Mistral 7B \\ {\tt layers.15.self\_attn.q\_proj}}
        \label{fig:param_heterogeneity:mistral7b}
    \end{subfigure}    \\
    \begin{subfigure}[b]{0.32\textwidth}
        \centering
        \includegraphics[width=\textwidth]{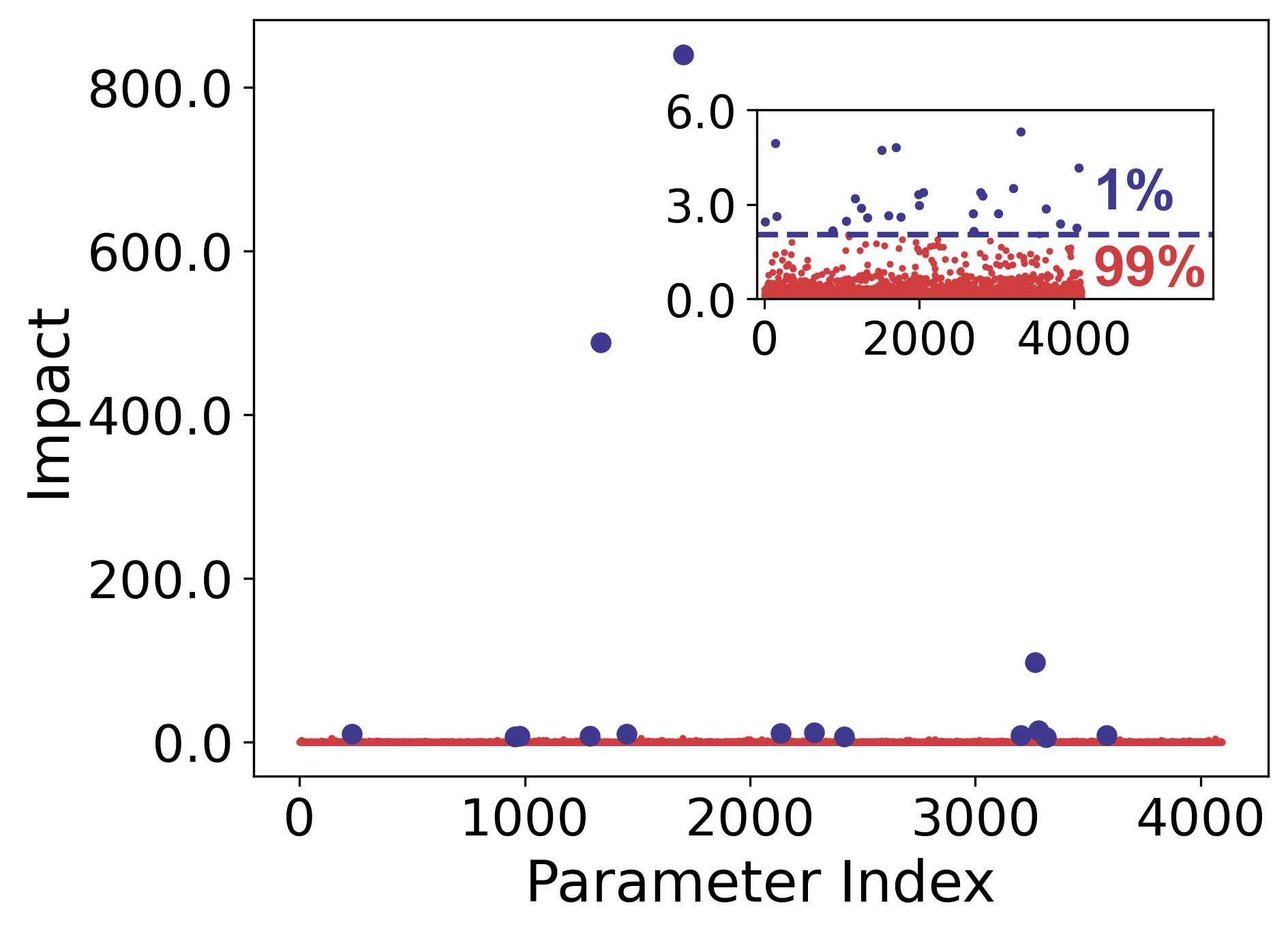}
        \caption{Gemma 7B \\ {\tt layers.25.mlp.down\_proj}}
        \label{fig:param_heterogeneity:gemma7b}
    \end{subfigure}
    \hfill
    \begin{subfigure}[b]{0.32\textwidth}
        \centering
        \includegraphics[width=\textwidth]{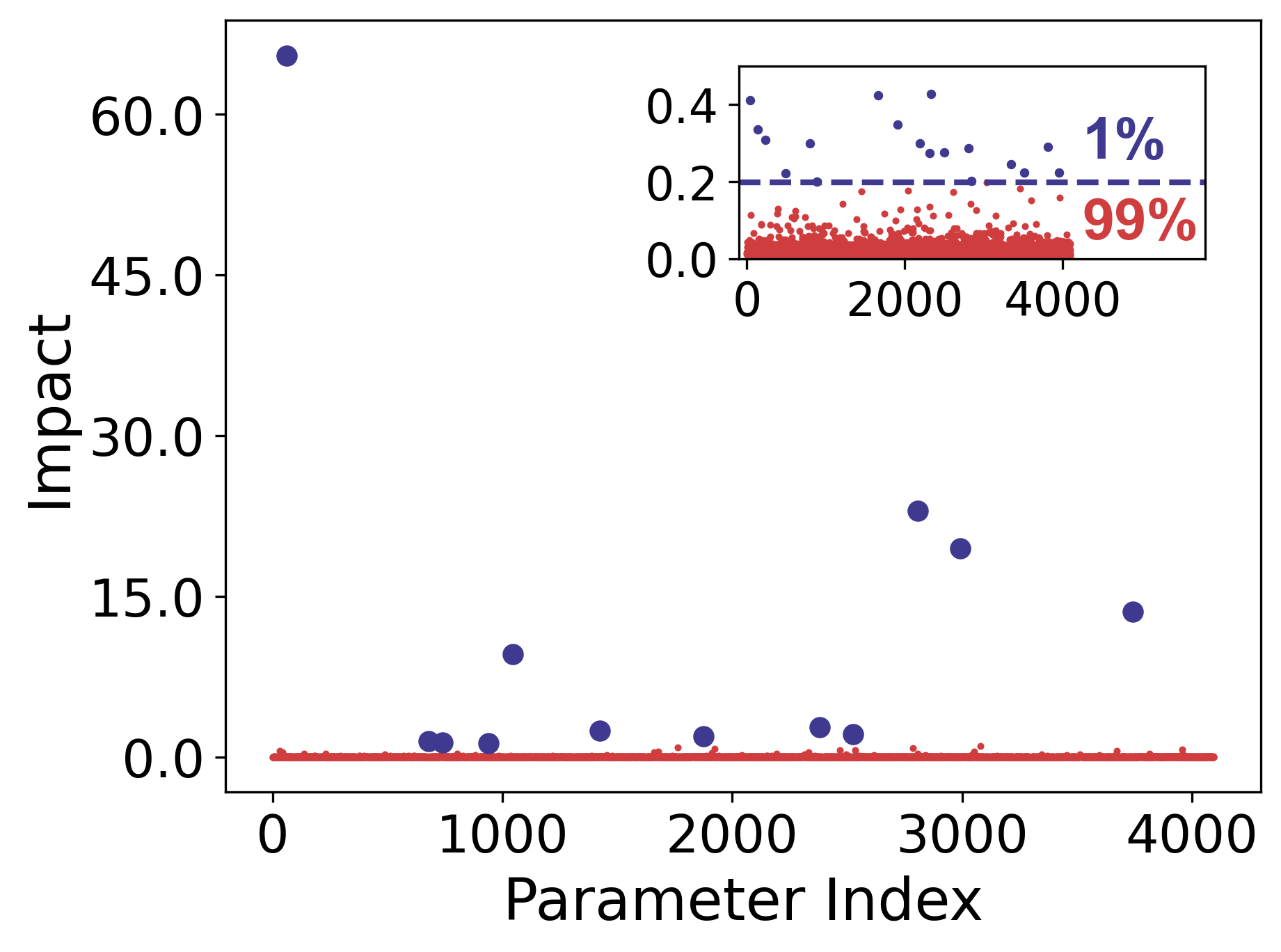}
        \caption{Vicuna-1.5 7B \\ {\tt layers.15.self\_attn.v\_proj}}
        \label{fig:param_heterogeneity:vicuna7b}
    \end{subfigure}
    \hfill
    \begin{subfigure}[b]{0.315\textwidth}
        \centering
        \includegraphics[width=\textwidth]{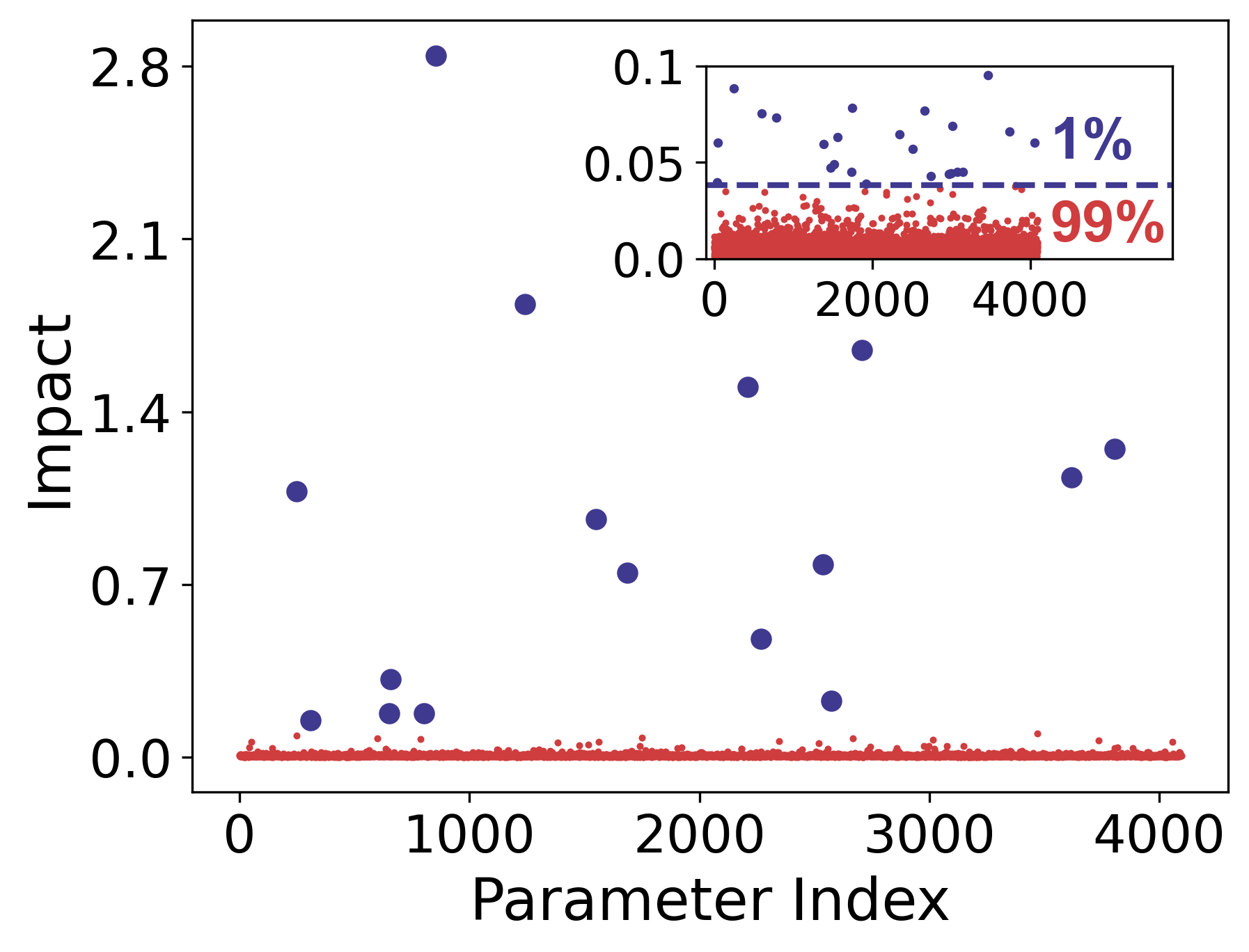}
        \caption{Vicuna-1.5 13B \\ {\tt layers.16.self\_attn.v\_proj}}
        \label{fig:param_heterogeneity:vicuna13b}
    \end{subfigure}    
    \caption{Scatter plot of parameter impacts in different LLMs. We randomly sampled 4096 parameters from the corresponding parameter matrix. Each point represents the impact of an individual parameter. Insets show the zoomed-in y-axis.  
    The heterogeneity is found across different model scales (\ref{fig:param_heterogeneity:llama7b},\ref{fig:param_heterogeneity:llama13b}), different model families (\ref{fig:param_heterogeneity:mistral7b}, \ref{fig:param_heterogeneity:gemma7b}), and both base models and chat models (\ref{fig:param_heterogeneity:vicuna7b}, \ref{fig:param_heterogeneity:vicuna13b}).}
    \label{fig:param_heterogeneity}
\end{figure}

Low-precision parameter representation leads to quantization errors. Surprisingly, existing research has shown that LLMs exhibit a high robustness for quantization errors even for low-bit settings. For example, although 4-bit quantization can only represent 16 distinct values, even the simplest round-to-nearest strategy does not significantly degrade performance~\cite{lin2023awq}. This raises the question: {\it what causes LLMs to be robust to quantization}?

We explore the parameters and answer this question via {\bf Parameter Heterogeneity}, which refers to the significant variation in the influence of quantization on different parameters. We reveal that for the vast majority ($>99\%$) of normal parameters, the effect of their quantization to the model are minimal and can thus be alleviated or ignored. However, there exists a small subset ($<1\%$) of ``cherry'' parameters for which the effect are substantial and hard to mitigate. %This parameter heterogeneity, where the impact of quantization is unevenly distributed among parameters, explains the high tolerance of LLMs to quantization.

Consider Figure~\ref{fig:param_heterogeneity:llama7b} as an example. We show a scatter plot of the impacts on the model loss when perturbing each individual parameter in a parameter matrix from LLaMA2-7b~\cite{touvron2023llama}. The derivation of impacts is detailed in \S~\ref{sec:measure}. While 99\% of the parameters fall within the range of (0, 0.1), a small subset of ``cherry'' parameters exhibits  values ranging from (5, 30), which is 50-300 times higher than the {\it maximum} value of the remaining 99\% of parameters. Moreover, this phenomenon is not an isolated case. We observed similar patterns across different scales of LLMs (Figure~\ref{fig:param_heterogeneity:llama7b}\ref{fig:param_heterogeneity:llama13b}), different families of LLMs, including Mistral~\cite{jiang2023mistral} (Figure~\ref{fig:param_heterogeneity:mistral7b}) and Gemma~\cite{team2024gemma} (Figure~\ref{fig:param_heterogeneity:gemma7b}), and both base models and chat models (Vicuna-1.5~\cite{chiang2023vicuna} Figure~\ref{fig:param_heterogeneity:vicuna7b}\ref{fig:param_heterogeneity:vicuna13b}). The consistent presence of this phenomenon suggests that parameter heterogeneity is an inherent characteristic of LLMs.

Therefore, 99\% of normal parameters explain the high robustness of LLMs to quantization errors. However, the small number of cherry parameters still leads to performance degradation under quantization. Consequently, the key to reducing quantization errors lies in addressing the quantization of cherry parameters.

The parameter heterogeneity also explains the previously discovered effectiveness of mixed-precision quantization strategies~\cite{dettmers2022gpt3,kim2023squeezellm,lin2023awq,liu2023llm}. By preserving a small proportion of parameters with high precision, the quantization performance can be effectively improved. Based on the heterogeneity, this strategy alleviates the impact of cherry parameters on model performance by maintaining their precision.

Indeed, mixed-precision strategies can effectively address the quantization error issue of cherry parameters. However, the core challenge lies in identifying cherry parameters based on specific metrics. Different metrics of parameter effects include weights~\cite{dettmers2022gpt3,kovaleva2021bert,liu2023llm}, activations~\cite{lin2023awq,wei2023outlier}, output changes~\cite{dettmers2023spqr}, as well as the impact of parameters on model loss used in this paper. Based on the parameter heterogeneity, we argue that effective cherry parameter identification metrics should exhibit high heterogeneity, clearly distinguishing between cherry parameters and normal parameters. Accordingly, we compare three different metrics in Sec~\ref{sec:phenomenon} and find that the impact best distinguishes cherry parameters from normal parameters. The experimental results in Sec~\ref{sec:exp:criteria} verify that choosing metrics with higher discriminative capability indeed leads to better performance.

Based on the above analysis, we design the CherryQ quantization algorithm, which selects cherry parameters based on the impact metric and end-to-end optimizes parameters with mixed precisions. Extensive experiments on various models and benchmarks demonstrate the effectiveness of CherryQ. It consistently yields the lowest perplexity in most settings. Notably, our 3-bit Vicuna-1.5 model exhibits performance on par with the 16-bit counterpart on Vicuna-bench~\cite{chiang2023vicuna}. Our 2-bit quantization method significantly outperforms the SOTA approaches.

In summary, by systematically revealing parameter heterogeneity in LLMs, we answer the following questions: 
\begin{enumerate}
    \item {\it What causes the high robustness of LLMs to quantization?} It is due to the 99\% of normal parameters in parameter heterogeneity.
    \item {\it Why is mixed-precision quantization effective?} This strategy addresses the quantization problem of the very few cherry parameters in parameter heterogeneity. 
    \item {\it How to find the optimal mixed-parameter selection strategy?} Based on heterogeneity that distinguishes normal parameters from cherry parameters, the impact-based metric demonstrates the highest discriminative capability.
    \item {\it How to quantize parameters according to parameter heterogeneity?} We propose CherryQ based on these findings. Extensive empirical results verify its effectiveness in 2-, 3-, 4-bit quantization.
\end{enumerate}

%% file: related.tex
\section{Related Work}

{\bf Quantization Strategies for LLMs} Various quantization strategies have been proposed in the literature to reduce the precision of weights and activations while maintaining acceptable accuracy. These strategies can be broadly classified into post-training quantization and quantization-aware training~\cite{krishnamoorthi2018quantizing}. Post-training quantization methods, such as OBD, OBS, and GPTQ, directly quantize the pre-trained model without fine-tuning~\cite{lecun1989optimal,hassibi1993optimal,frantar2023gptq}. On the other hand, quantization-aware training methods, such as LLM-QAT~\cite{liu2023llm}, incorporate quantization operations into the training process to jointly optimize the quantized model. Some works also explore mixed-precision quantization~\cite{kim2023squeezellm} and adaptive quantization bins~\cite{dettmers2024qlora} to achieve a better trade-off between accuracy and efficiency.

{\bf Outliers in Language Model Quantization} 
The idea of modeling parameter outliers in LLM quantization is not new. Exploring outliers mainly includes the perspectives of magnitude~\cite{liu2023llm,dettmers2024qlora} and activations~\cite{bondarenko2021understanding,dettmers2022gpt3}. For example, from the magnitude perspective, QLoRA assumes that parameters follow a Gaussian distribution~\cite{dettmers2024qlora} and designs information-theoretically optimal quantized bins based on this assumption. \cite{liu2023llm} keeps outlier parameters in 16-bit precision. From the activation perspective, \cite{lin2023awq} migrates the outlier amplifier to subsequent modules through an equivalent transformation. Additionally, SqueezeLLM also measures outliers from the perspective of parameter impact~\cite{kim2023squeezellm}. To the best of our knowledge, our work is the first to systematically reveal the outliers (heterogeneity) of parameter impact across different models, and we show a more pronounced imbalance in parameter impacts compared to magnitudes (\S~\ref{sec:exp:criteria}). Furthermore, we propose a method to unify outlier (cherry) parameter optimization and normal parameter optimization, addressing the optimization challenges of heterogeneous parameters.

%% file: measure.tex
\section{Quantifying the Impact of Parameters on Model Performance}
\label{sec:measure}

The impact of parameters on model performance is quantified by the increase of the training loss when perturbing the parameter weight, which is widely used in post-training quantization approaches~\cite{lecun1989optimal,hassibi1993optimal,frantar2023gptq}. We adopt a second-order Taylor approximation of the training loss w.r.t. parameter perturbation. Given a parameter $w_i$ and a small perturbation $\Delta$ applied to it, such that $w_i \leftarrow w_i + \Delta$, the change in the training loss can be expressed as:
\begin{equation}
\mathbf{L}(w_i+\Delta) - \mathbf{L}(w_i) = g_i\Delta+\frac{1}{2} \mathbf{H}_{ii} \Delta^2 + O(\Delta^2)
\end{equation}
where $g_i=\mathbb{E}[\frac{\partial L}{\partial w_i}]$ represents the expected gradient of the loss with respect to $w_i$, and $\mathbf{H}_{ii}=\mathbb{E}[\frac{\partial^2 L}{\partial w_i^2}]$ denotes the $i$-th value of the Hessian matrix of the loss. Since the target model is a well-converged model, we can assume that $g_i \approx 0$, simplifying the expression to:
\begin{equation}
    \mathbf{L}(w_i+\Delta) - \mathbf{L}(w_i) \approx \frac{1}{2}\mathbf{H}_{ii} \Delta^2
\end{equation}
\iffalse
The quantization operation can be represented as $w_i +\Delta = quant(w_i)$, where $quant(\cdot)$ denotes the quantization function. Substituting this into the above expression yields:
\begin{equation}
\mathbf{L}(w_i+\Delta) - \mathbf{L}(w_i) \approx \frac{1}{2}(quant(w_i)-w_i)^T \mathbf{H}_i (quant(w_i)-w_i)    
\end{equation}
\fi

Therefore, $\mathbf{H}_{ii}$ quantify the impact of quantization-induced perturbations on the model's training loss. Parameters with larger values of $\mathbf{H}_{ii}$ exhibit higher sensitivity to quantization and require careful treatment to maintain model performance. We denote $\mathbf{H}_{ii}$ as the impact of $w_i$.

{\bf Efficient Computation}
Computing $\mathbf{H}_{ii}$ of the diagonal of Hessian matrix for each parameter is computationally expensive, particularly for large-scale models. To overcome this challenge, we propose an efficient approximation using the Fisher Information Matrix ($\mathbf{F}$). Since $\mathbf{H}$ is the Hessian matrix of a negative log-likelihood loss, $\mathbf{H}$ is equal to Fisher information matrix~\cite{li2020brecq}. For the diagonal of the Hessian matrix, we have:
% , defined as the expected outer product of the gradients:
\begin{equation}
 \mathbf{H}_{ii} = \mathbf{F}_{ii} = \mathbb{E}[g_i ^2]   
\end{equation}
% The diagonal of the FIM serves as an approximation to the diagonal of the Hessian matrix, allowing for efficient estimation of the impact of parameter perturbations. Modern deep learning frameworks enable the computation of the FIM with minimal overhead during the training process. 

%% file: method.tex
\section{End-to-End Mixed-Precision Quantization}

The insights gained from Figure~\ref{fig:param_heterogeneity} highlight the heterogeneity in model parameters. %The cherry parameters, despite constituting less than 1\% of the total parameter count, exert a substantial influence on the model. Indiscriminately quantizing these cherry parameters alongside the normal parameters may lead to a significant deterioration in model performance.
To mitigate the impact of cherry parameters on quantization, we propose to preserve their high-precision values during the quantization process. By maintaining the fidelity of these critical parameters, we ensure that the essential information they capture is not compromised. 

%Optimizing mixed-precision parameters in LLMs presents a unique challenge. The widely adopted GPTQ approach~\cite{frantar2023gptq}, which falls under the Post-Training Quantization (PTQ) framework~\cite{krishnamoorthi2018quantizing}, struggles to simultaneously optimize high-precision cherry parameters and low-precision normal parameters. This is because updating the cherry parameters during the PTQ process significantly affects the model, causing the optimal values of the normal parameters to vary. However, in the PTQ framework, once the parameters are quantized, they cannot be updated further. This limitation prevents the early-stage quantized parameters from reaching their optimal values. On the other hand, 

Optimizing mixed-precision parameters in LLMs presents a unique challenge in the widely adopted Post-Training Quantization (PTQ) framework~\cite{krishnamoorthi2018quantizing}. If we do not allow the updates of the cherry parameters, the quantization will certainly lose the flexibility provided by these critical parameters. This prevents the cherry parameters from reaching their optimum. On the other hand, PTQ struggles to simultaneously optimize high-precision cherry parameters and low-precision normal parameters. This is because the cherry parameter updates during the PTQ process significantly affect the optimal values of the normal parameters. So normal parameters need to be continually updated as the cherry parameter varies. However, in PTQ, once the normal parameters are quantized, they cannot be further updated. This prevents the early-stage quantized parameters from reaching their optimal values.

To address this challenge, we propose a novel approach that end-to-end optimize the mixed-precision parameters via backpropagation. Our method leverages a quantization-aware training framework. To simultaneously optimize both the cherry parameters and normal parameters, we use two separate backpropagation strategies. The high-precision cherry parameters are updated using standard gradient descent, while the low-precision normal parameters employ the Straight-Through Estimator (STE) trick~\cite{bengio2013estimating} for low-precision gradient descent. This unified backpropagation enables the end-to-end optimization of both cherry parameters and normal parameters, enhancing the overall optimization effect. We show the quantization in Algorithm~\ref{algo}.

\begin{algorithm}[h]
\caption{CherryQ}
\begin{algorithmic}[1]
\Require Model parameters $\mathbf{W}$, quantization function $Quant(\cdot)$, threshold $\tau$, learning rate $\eta$
\Ensure Quantized model parameters

\State $\mathbf{C} \gets \{w_i \in \mathbf{W} \mid \mathbf{H}_{ii} > \tau\}$ \Comment{Identify cherry parameters}
\State $\mathbf{N} \gets \mathbf{W} \setminus \mathbf{C}$ \Comment{Identify normal parameters}

\For{each training batch $x$}
\State $L \gets \mathrm{model}(x;\mathbf{C} \cup Quant(\mathbf{N}))$ \Comment{Compute loss w.r.t. mixed-precision parameters}

\State $\mathbf{C} \gets \mathbf{C} - \eta \frac{\partial L}{\partial \mathbf{C}}$ \Comment{Standard gradient descent for cherry parameters}
\State $\mathbf{N} \gets \mathbf{N} - \eta \cdot \mathrm{STE}(\frac{\partial L}{\partial \mathbf{N}})$ \Comment{Gradient approximation by STE for normal parameters}
\EndFor

\State \textbf{return} $\mathbf{C} \cup Quant(\mathbf{N})$
\end{algorithmic}
\label{algo}
\end{algorithm}

%% file: phenomenon.tex
\section{Heterogeneity-based Cherry Parameter Selection}
\label{sec:phenomenon}

Correctly identifying cherry parameters is one of the main challenges of CherryQ quantization. Candidate metrics for parameter influences include weights~\cite{kovaleva2021bert,liu2023llm,dettmers2022gpt3}, activations~\cite{lin2023awq,wei2023outlier}, and impacts ($\mathbf{H}_{ii}$). We propose that an effective metric should reflect heterogeneity, specifically by differentiating the influence of cherry parameters and normal parameters of the model.

To this end, we define the heterogeneity score. In Figure~\ref{fig:param_heterogeneity}, a small subset of parameters exhibit significantly higher impacts compared to the maximum of the majority. Inspired by this, the heterogeneity score is defined as the ratio of the {\it mean} impact of the top 1\% parameters to the {\it maximum} impact of the bottom 99\% parameters, as shown in Equation~\eqref{eq:heterogeneity_score}. A higher heterogeneity score indicates a more significant disparity in parameter importance.

\begin{equation}
\text{Heterogeneity Score}(f) = \frac{\text{Mean}(f(w_i)_{\text{ top 1\%}})}{\text{Max}(f(w_i)_{\text{ bottom 99\%}})}
\label{eq:heterogeneity_score}
\end{equation}
where $f(w_i)$ denotes the parameter influence for parameter $w_i$, and $f$ is chosen from parameter weights, activations, and impacts.

\begin{figure}[htbp]
    \centering
    \begin{subfigure}[b]{0.32\textwidth}
        \centering
        \includegraphics[width=\textwidth]{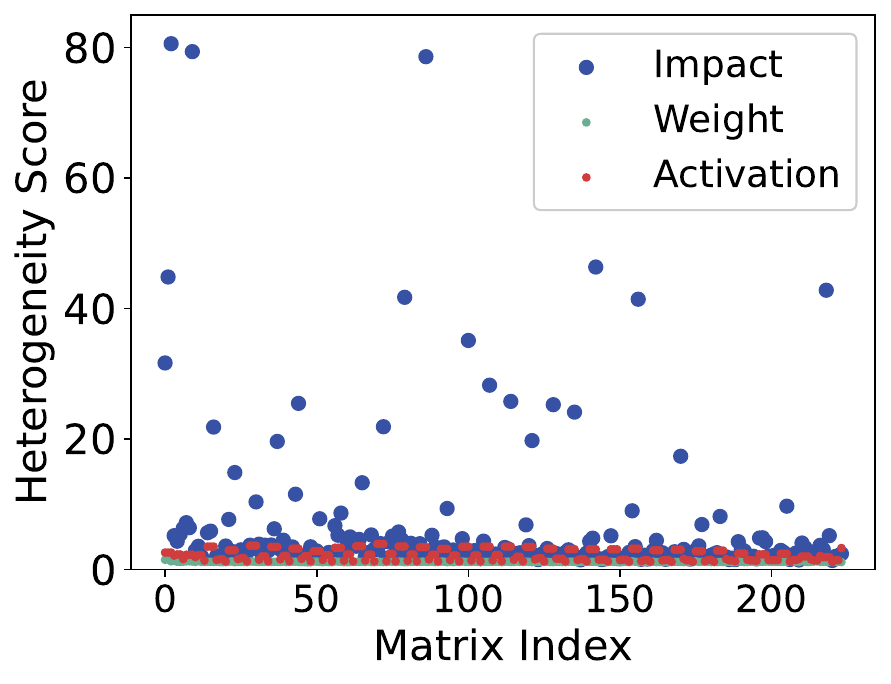}
        \caption{LLaMA-2 7B}
    \end{subfigure}
    \hfill
    \begin{subfigure}[b]{0.33\textwidth}
        \centering
        \includegraphics[width=\textwidth]{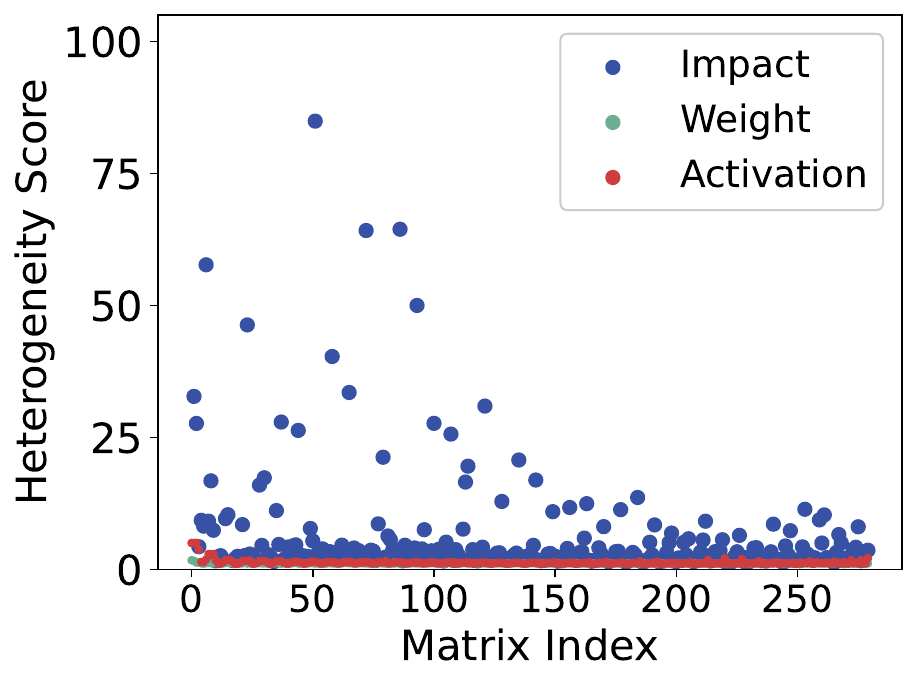}
        \caption{LLaMA-2 13B}
    \end{subfigure}
    \hfill
    \begin{subfigure}[b]{0.32\textwidth}
        \centering
        \includegraphics[width=\textwidth]{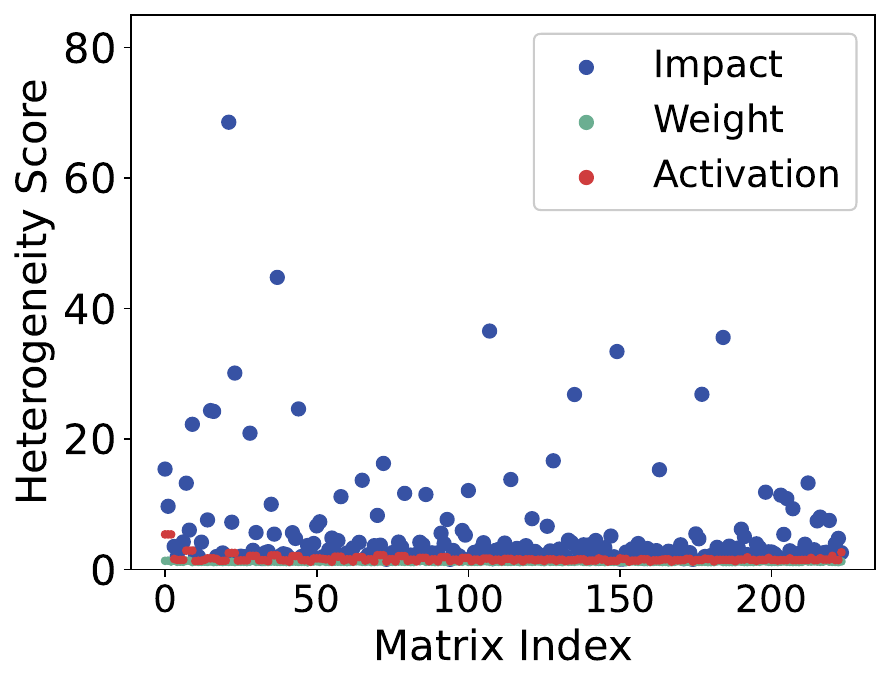}
        \caption{Mistral 7B}
    \end{subfigure}
    
    \begin{subfigure}[b]{0.33\textwidth}
        \centering
        \includegraphics[width=\textwidth]{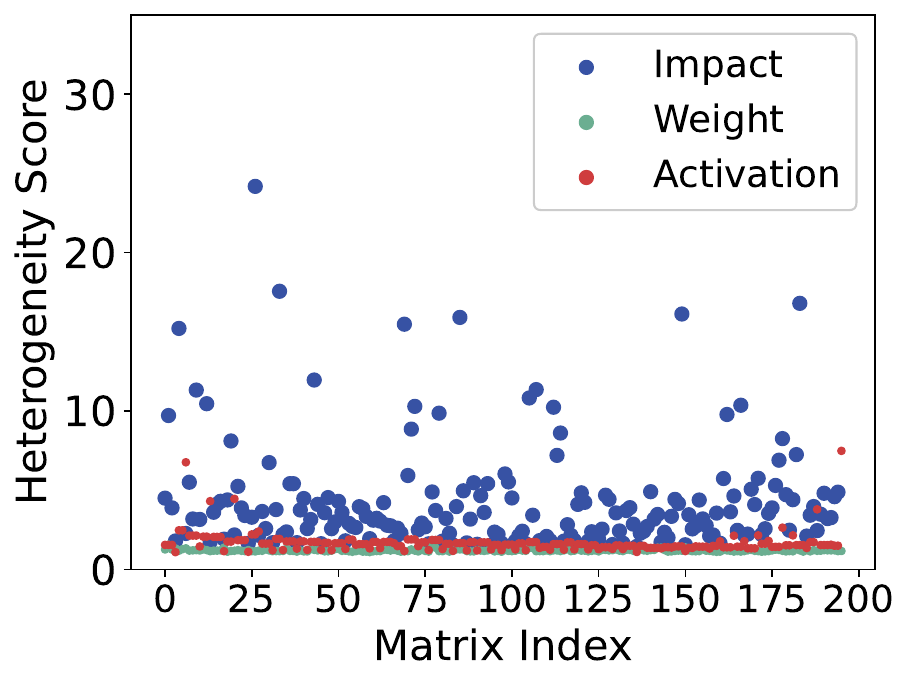}
        \caption{Gemma 7B}
    \end{subfigure}
    \hfill
    \begin{subfigure}[b]{0.32\textwidth}
        \centering
        \includegraphics[width=\textwidth]{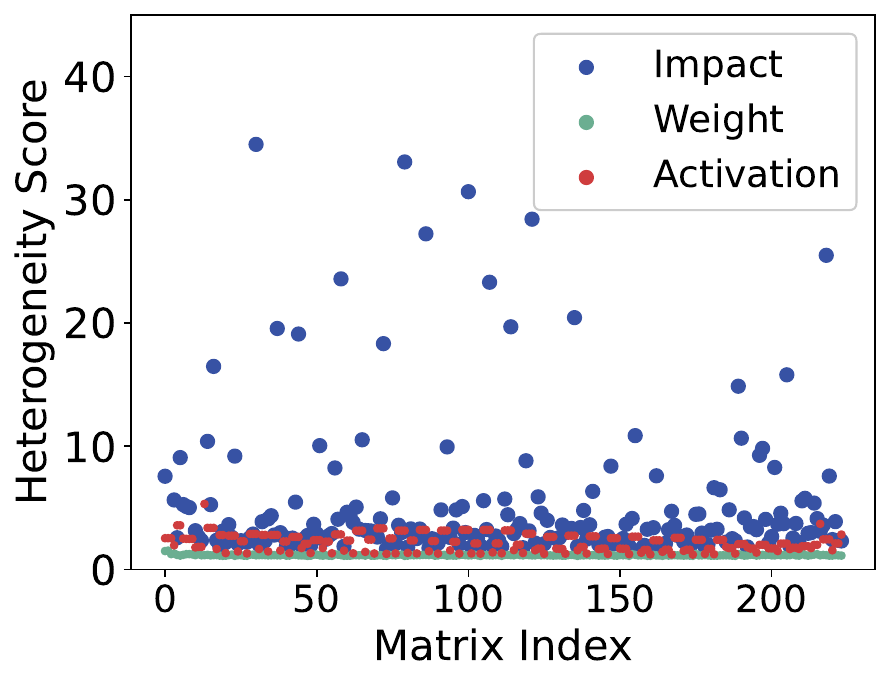}
        \caption{Vicuna-1.5 7B}
    \end{subfigure}
    \hfill
    \begin{subfigure}[b]{0.32\textwidth}
        \centering
        \includegraphics[width=\textwidth]{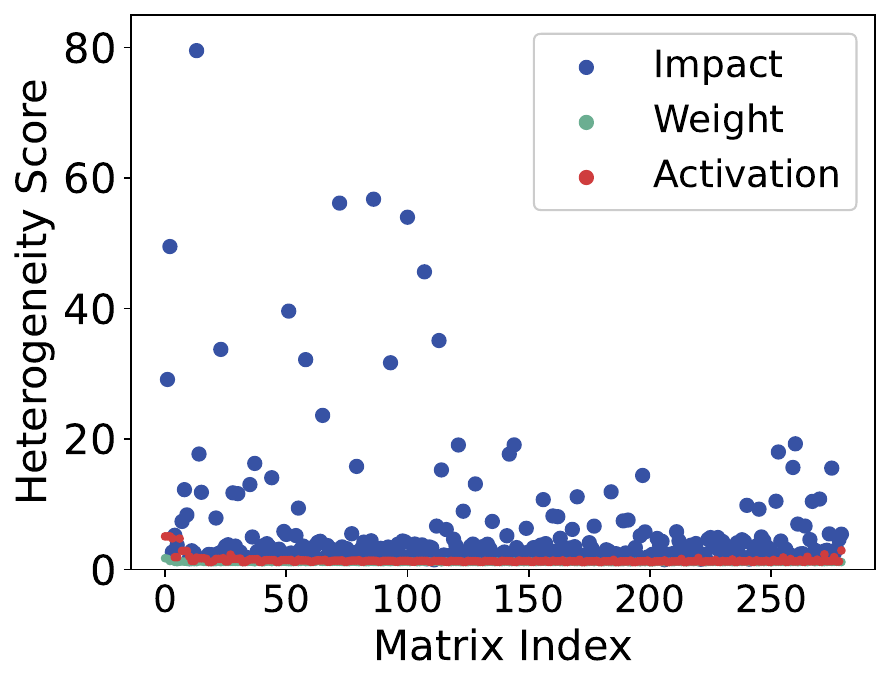}
        \caption{Vicuna-1.5 13B}
    \end{subfigure}
    
    \caption{Scatter distribution of heterogeneity scores for different parameter matrices in LLMs. Each point represents a parameter matrix.}
    \label{fig:heterogeneity_scatter}
\end{figure}

Figure~\ref{fig:heterogeneity_scatter} presents the heterogeneity scores for different metrics across various LLMs. The impact-based metric consistently shows higher heterogeneity scores compared to weights and activations. This indicates that the impact metric better distinguishes between the normal and cherry parameters, thus providing a more effective means of identifying cherry parameters. The validity of using heterogeneity scores for cherry parameter selection will be further verified in Sec~\ref{sec:exp:criteria}, demonstrating that higher heterogeneity scores lead to better model performance.

%The superiority of the impact-based metric is further validated by the experimental results in Sec~\ref{sec:exp:criteria}, where models quantized using cherry parameters identified by the impact metric achieve the lowest perplexity scores.

{\bf Data Independence} We investigate whether impact-based parameter heterogeneity exhibits data independence - that is, whether different data samples share the same cherry parameters. Applying identical cherry parameters across different samples is only valid when there is data independence. To examine data independence within the same dataset, we randomly selected five sets of 128 WikiText-2 samples. We evaluated the overlap of their cherry parameters in each pair of sample sets. To evaluate cross-dataset data independence, we performed 9 independent sampling trials, each collecting 128 samples from C4 and 128 samples from WikiText-2 to evaluate the overlap. For Vicuna, we further added Sharegpt as a data source. Table~\ref{tab:data_independence} presents the overlap ratios of the cherry parameters. Despite the fact that cherry parameters constitute only $1/256$ of the total parameters, all models demonstrate significant overlap ratios. This finding suggests that the cherry parameters possess an inherent data-independent nature.

\begin{table}[t]
\centering
\caption{Overlap ratio (\%) of cherry parameters (top 1/256), where L2: LLaMA-2, V: Vicuna-1.5, M: Mistral, G: Gemma.}
\begin{tabular}{lcccccc}
\toprule
\textbf{Model} & \textbf{L2-7B} & \textbf{L2-13B} & \textbf{V-7B} & \textbf{V-13B} & \textbf{M-7B} & \textbf{G-7B} \\
\midrule
Within dataset & 84 & 75 & 68 & 63 & 89 & 90 \\
Across datasets & 68 & 66 & 65 & 60 & 85 & 86 \\
\bottomrule
\end{tabular}
\label{tab:data_independence}
\end{table}

%% file: experiments.tex
\section{Quantization Experiments}

%In the experimental section, we demonstrate the effectiveness of the CherryQ for both base LLMs and chat LLMs. We also compare different cherry parameter selection criteria to highlight the effect of impact-based heterogeneity.

\subsection{Implementation Details}
\label{sec:exp:implement_details}

%All experiments were completed on a single node with 8*A100 80G GPUs.

\textbf{Parameter Representation:} Based on the observation that cherry parameters occupy a very small proportion, for each row of parameters in each parameter matrix, we consider only the top 1/256 parameters with the highest impact as cherry parameters and retain their FP16 precisions. For example, the parameter matrix size of LLaMA2-7B is $4096 \times 4096$. So we average the impact across all rows for each column and then select the top $16$ columns with the highest average impact, resulting in $16 \times 4096$ parameters as cherry parameters. Furthermore, to recover the complete parameter matrix, an INT16 is required to record the indices of these $16$ columns. Thus, the storage overhead for the column indices is minimal. For normal parameters, we employ {\it full range symmetric MinMax quantization} to quantize their weights. And we adopt a widely-used parameter grouping strategy. For more details, see Sec~\ref{sec:append:quant}.

%\subsection{Experimental Setup}

%\textbf{Baselines:} We compare our method with various quantization methods, including QAT~\cite{liu2023llm}, GPTQ~\cite{frantar2023gptq}, SqueezeLLM~\cite{kim2023squeezellm}, OmniQuant~\cite{shao2023omniquant}, and AWQ~\cite{lin2023awq}. More details of the baselines are shown in \S~\ref{sec:append:setup}.

%The 4-bit GPTQ model is obtained from the open-source community~\footnote{https://huggingface.co/TheBloke}. Due to the lack of a 3-bit GPTQ model, we quantize the model ourselves via Auto-GPTQ~\footnote{https://github.com/AutoGPTQ/AutoGPTQ}. We implement the QAT quantization ourselves. 

%For a fair comparison, we do not use knowledge distillation as in LLM-QAT~\cite{liu2023llm}. All settings are equivalent for CherryQ and QAT, \textcolor{red}{except for the absence of additional processing of cherry parameters, the implementations of QAT are consistent with CherryQ.} 

%The perplexity baselines of OmniQuant and AWQ from \cite{shao2023omniquant}. 
% It is suggested to create a seperate Chapter named "Evaluation" to explain the evaluation details.
%All quantization methods use LLaMA2 as the base model for quantization and Vicuna 1.5 as the chat model for quantization.

\textbf{Quantization Datasets:} For the quantization of the base LLMs, we follow~\cite{frantar2023gptq} to use C4~\cite{raffel2020exploring} as the training data. We selected the first four partitions of C4 and chose data with a length of $\geq 2048$ tokens, resulting in a total of 50k samples of 2048 tokens. 
%For GPTQ, we follow~\cite{frantar2023gptq} to use 128 randomly selected 2048-token segments from the first partition of C4.
For the chat LLMs, since Vicuna-1.5~\cite{chiang2023vicuna} is obtained by supervised fine-tuning based on ShareGPT~\cite{chiang2023vicuna}, we also use the ShareGPT dataset for training. We used a total of 20k training samples from ShareGPT for QAT and Cherry. 
%For GPTQ, We follow~\cite{frantar2023gptq} to use 128 randomly selected 2048-token segments from the 20k samples as the calibration data.

\textbf{Baselines} We compare our method with various quantization methods, including QAT~\cite{liu2023llm}, GPTQ~\cite{frantar2023gptq}, SqueezeLLM~\cite{kim2023squeezellm}, OmniQuant~\cite{shao2023omniquant}, and AWQ~\cite{lin2023awq}.
For OmniQuant and AWQ, we use their results reported in~\cite{shao2023omniquant}. For SqueezeLLM, we use the results in its original paper~\cite{kim2023squeezellm}. For GPTQ, its 4-bit model is obtained from the open-source~\footnote{https://huggingface.co/TheBloke}. Due to the lack of a 3-bit GPTQ model, we quantize the model ourselves via the implementation of Auto-GPTQ~\footnote{https://github.com/AutoGPTQ/AutoGPTQ}. Since CherryQ is based on QAT, for fair comparisons, the implementation of QAT is the same as CherryQ, except that it does not handle cherry parameters.

% We select the first 4 partitions of C4 and choose data with a length of $\geq 2048$ tokens, resulting in a total of 50k samples.
%\textcolor{red}{Except for the calibration dataset of GPTQ consisting of 128 randomly selected 2048-token segments from the first partition of C4, we select the first 4 partitions of C4 and choose data with a length of $\geq 2048$ tokens, resulting in a total of 50k samples of 2048 token.}
%For the quantization of the chat LLMs, since Vicuna-1.5~\cite{chiang2023vicuna} is obtained by supervised fine-tuning based on ShareGPT~\cite{chiang2023vicuna}, we also use the ShareGPT dataset for quantization. We utilize a total of 20k ShareGPT data samples. \textcolor{red}{To negate data bias, we randomly select 128 samples from those 20k samples as the chat calibration dataset of GPTQ.}

%More details of the setup are shown in \S~\ref{sec:append:setup}.

\subsection{Effect of Base LLM Quantization}

In this section, we present the main experimental results to demonstrate the effectiveness of CherryQ on LLaMA2~\cite{touvron2023llama}. We evaluate CherryQ with both perplexity and downstream tasks, comparing its performance with state-of-the-art quantization methods.

\subsubsection{Perplexity Results}

We follow~\cite{frantar2023gptq,shao2023omniquant} to evaluate the perplexity of CherryQ on two widely used corpora: C4 and WikiText-2~\cite{merity2016pointer}. We use the validation split of C4 to avoid data leakage. We show the results of 3-bit quantization using different quantization approaches in Table~\ref{tab:model_comparison_3bit}. We show the results of different model scales and different group sizes.

From the results, CherryQ consistently outperforms all other approaches across both model sizes (7B and 13B) and grouping sizes (64 and 128), achieving the lowest perplexity on both the C4 and WikiText-2 datasets. Notably, CherryQ's perplexity is significantly closer to the full-precision (FP16) baseline compared to other methods, highlighting its ability to preserve model performance after quantization.

Table~\ref{tab:model_comparison_4bit} compares different 4-bit quantization methods. Again, CherryQ achieves the lowest perplexity scores in most settings, demonstrating its effectiveness in higher-bit quantization settings.

\subsubsection{Downstream Task Performance}

To further validate the effectiveness on specific tasks, we evaluated the quantized models on various downstream tasks from the HuggingFace OpenLLM Leaderboard. 
Table~\ref{tab:quantization_comparison_transposed_formatted} presents the performance comparison of different 3-bit quantization methods for LLaMA2. {\bf CherryQ consistently outperforms other methods across almost all tasks}, achieving the highest average score. This showcases CherryQ's ability to maintain the model's generalization capabilities for downstream tasks.

% \begin{table}[h]
% \centering
% \caption{Perplexity (\textcolor{red}{$\downarrow$}) of 3-bit quantization on LLaMA2 models . gX means the group size is X. The results of OmniQuant and AWQ are from~\cite{shao2023omniquant}. The results of SqueezeLLM are from~\cite{kim2023squeezellm}.}
% \label{tab:model_comparison_3bit}
% \begin{tabular}{lcccccccc}
% \toprule
% Method  & \multicolumn{2}{c}{7B-3bit-g128} & \multicolumn{2}{c}{7B-3bit-g64} & \multicolumn{2}{c}{13B-3bit-g128} & \multicolumn{2}{c}{13B-3bit-g64} \\
% \cmidrule(lr){2-3} \cmidrule(lr){4-5} \cmidrule(lr){6-7} \cmidrule(lr){8-9}
% & c4 & wiki2 & c4 & wiki2 & c4 & wiki2 & c4 & wiki2 \\
% \midrule
% FP16& 6.97 & 5.47 & 6.97 & 5.47 & 6.47 & 4.88 & 6.47 & 4.88 \\
% QAT  & 9.25 & 6.90 & 8.74 & 7.13 & 7.19 & 5.63 & 7.02 & 5.48 \\
% GPTQ & 8.28 & 6.74 & 8.20 & 6.62 & 7.24 & 5.63 & 7.10 & 5.56 \\
% OmniQuant & 7.75 & 6.03 & - & - & 6.98 & 5.28 & - & - \\
% AWQ & 7.84 & 6.24 & - & - & 6.94 & 5.32 & - & - \\
% SqueezeLLM  & 7.51 & 5.96 & - & -  & 6.82 & {\bf 5.23} & - & - \\
% \cellcolor{highlight}\textbf{CherryQ} & \cellcolor{highlight}\textbf{7.39} & \cellcolor{highlight}\textbf{5.93} & \cellcolor{highlight}\textbf{7.34} & \cellcolor{highlight}\textbf{5.87} & \cellcolor{highlight}\textbf{6.80} & \cellcolor{highlight}5.26 & \cellcolor{highlight}\textbf{6.76} & \cellcolor{highlight}\textbf{5.21} \\
% \bottomrule
% \end{tabular}
% \end{table}

\begin{table}[h]
\centering
\small
\setlength{\tabcolsep}{4pt}
\caption{Perplexity (\textcolor{red}{$\downarrow$}) of 3-bit quantization on LLaMA2 models. gX means the group size is X. The results of OmniQuant and AWQ are from~\cite{shao2023omniquant}. The results of SqueezeLLM are from~\cite{kim2023squeezellm}.}
\label{tab:model_comparison_3bit}
\begin{tabular}{lcccccccccccc}
\toprule
Method & \multirow{2}{*}{\shortstack{Avg.\\bit}} & \multicolumn{2}{c}{7B-3bit-g128} & \multirow{2}{*}{\shortstack{Avg.\\bit}} & \multicolumn{2}{c}{7B-3bit-g64} & \multirow{2}{*}{\shortstack{Avg.\\bit}} & \multicolumn{2}{c}{13B-3bit-g128} & \multirow{2}{*}{\shortstack{Avg.\\bit}} & \multicolumn{2}{c}{13B-3bit-g64} \\
\cmidrule(lr){3-4} \cmidrule(lr){6-7} \cmidrule(lr){9-10} \cmidrule(lr){12-13}
& & c4 & wiki2 & & c4 & wiki2 & & c4 & wiki2 & & c4 & wiki2 \\
\midrule
FP16& 16 & 6.97 & 5.47 & 16 & 6.97 & 5.47 & 16 & 6.47 & 4.88 & 16 & 6.47 & 4.88 \\
QAT  & 3.13 & 9.25 & 6.90 & 3.25 & 8.74 & 7.13 & 3.13 & 7.19 & 5.63 & 3.25 & 7.02 & 5.48 \\
GPTQ & 3.15 & 8.28 & 6.74 & 3.30 & 8.20 & 6.62 & 3.15 & 7.24 & 5.63 & 3.30 & 7.10 & 5.56 \\
AWQ & 3.15 & 7.84 & 6.24 & - & - & - & 3.15 & 6.94 & 5.32 & - & - & - \\
OmniQuant & 3.15 & 7.75 & 6.03 & - & - & - & 3.15 & 6.98 & 5.28 & - & - & - \\
SqueezeLLM  & - & - & - & 3.24 & 7.51 & 5.96  & - & - & - & 3.24 & 6.82 & 5.23 \\
\cellcolor{highlight}\textbf{CherryQ} & \cellcolor{highlight}3.17 & \cellcolor{highlight}\textbf{7.39} & \cellcolor{highlight}\textbf{5.93} & \cellcolor{highlight}3.30 & \cellcolor{highlight}\textbf{7.34} & \cellcolor{highlight}\textbf{5.87} & \cellcolor{highlight}3.17 & \cellcolor{highlight}\textbf{6.80} & \cellcolor{highlight}\textbf{5.26} & \cellcolor{highlight}3.29 & \cellcolor{highlight}\textbf{6.76} & \cellcolor{highlight}\textbf{5.21} \\
\bottomrule
\end{tabular}
\end{table}

Table~\ref{tab:quantization_comparison_4bit_adjusted} extends the comparison to 4-bit quantization. CherryQ continues to excel, achieving the highest scores on most individual tasks and the highest average score overall. These results highlight the generalization ability of CherryQ across different quantization bits and model sizes.

% \begin{table}[h]
% \centering
% \caption{Perplexity (\textcolor{red}{$\downarrow$}) of 4-bit quantization on LLaMA2 models.}
% \label{tab:model_comparison_4bit}
% \begin{tabular}{lcccc}
% \toprule
% Method & \multicolumn{2}{c}{7B-4bit-g128} & \multicolumn{2}{c}{13B-4bit-g128} \\
% \cmidrule(lr){2-3} \cmidrule(lr){4-5}
% & c4 & wiki2 & c4 & wiki2 \\
% \midrule
% FP16 & 6.97 & 5.47 & 6.47 & 4.88 \\
% QAT & 7.29 & 5.81 & 6.67 & 5.12 \\
% GPTQ & 7.30 & 5.73 & 6.63 & 4.97 \\
% OmniQuant & 7.12 & 5.58 & 6.56 & \textbf{4.95} \\
% AWQ & 7.13 & 5.62 & 6.56 & 4.97 \\
% %SqLLM & 7.08 & 5.57 & 6.54 & 4.96 \\
% \rowcolor{highlight}\textbf{CherryQ} & \textbf{7.07} & \textbf{5.58} & \textbf{6.56} & 4.99 \\
% \bottomrule
% \end{tabular}
% \end{table}

\begin{table}[h]
\centering
\caption{Perplexity (\textcolor{red}{$\downarrow$}) of 4-bit quantization on LLaMA2 models.}
\label{tab:model_comparison_4bit}
\begin{tabular}{lcccccc}
\toprule
Method & \multirow{2}{*}{\shortstack{Avg.\\bit}} & \multicolumn{2}{c}{7B-4bit-g128} & \multirow{2}{*}{\shortstack{Avg.\\bit}} & \multicolumn{2}{c}{13B-4bit-g128} \\
\cmidrule(lr){3-4} \cmidrule(lr){6-7}
& & c4 & wiki2 & & c4 & wiki2 \\
\midrule
FP16 & 16 & 6.97 & 5.47 & 16 & 6.47 & 4.88 \\
QAT & 4.13 & 7.29 & 5.81 & 4.13 & 6.67 & 5.12 \\
GPTQ & 4.15 & 7.30 & 5.73 & 4.15 & 6.63 & 4.97 \\
AWQ & 4.15 & 7.13 & 5.62 & 4.15 & 6.56 & 4.97 \\
OmniQuant & 4.15 & 7.12 & 5.58 & 4.15 & 6.56 & \textbf{4.95} \\
%SqLLM & 7.08 & 5.57 & 6.54 & 4.96 \\
\rowcolor{highlight}\textbf{CherryQ} & 4.17 & \textbf{7.07} & \textbf{5.58} & 4.16 & \textbf{6.56} & 4.99 \\
\bottomrule
\end{tabular}
\end{table}

\begin{table}[h]
\centering
\caption{Performance of different 3-bit quantization methods on Huggingface OpenLLM for LLaMA2-7B and
LLaMA2-13B.}
\label{tab:quantization_comparison_transposed_formatted}
\begin{tabular}{l|cccccc|c}
\toprule
Method & Hellaswag & Winogrande & ARC & TruthfulQA & GSM8K & MMLU & Average (\textcolor{red}{$\uparrow$}) \\
\midrule
\multicolumn{8}{c}{LLaMA2-7B-3bit-g64} \\
\midrule
FP16 & 78.6 & 74.0 & 53.2 & 38.8 & 14.5 & 46.7 & 51.0 \\
QAT & 75.5 & 71.6 & 49.2 & 37.3 & 7.3 & 40.6 & 46.9 \\
GPTQ & 73.9 & 71.7 & 48.6 & 38.8 & 8.1 & 39.4 & 46.8 \\
\cellcolor{highlight}\textbf{CherryQ} & \cellcolor{highlight}{\bf 77.0} & \cellcolor{highlight}{\bf 71.8} & \cellcolor{highlight}{\bf 50.6} & \cellcolor{highlight}{\bf 38.6} & \cellcolor{highlight}{\bf 10.4} & \cellcolor{highlight}{\bf 43.9} & \cellcolor{highlight}\textbf{48.7} \\
\midrule
\multicolumn{8}{c}{LLaMA2-7B-3bit-g128} \\
\midrule
FP16 & 78.6 & 74.0 & 53.2 & 38.8 & 14.5 & 46.7 & 51.0 \\
QAT & 75.4 & 70.8 & 48.2 & 37.7 & 6.7 & 39.0 & 46.3 \\
GPTQ & 72.9 & 70.8 & 48.6 & 39.1 & 5.4 & 38.2 & 45.8 \\
\cellcolor{highlight}\textbf{CherryQ} & \cellcolor{highlight}{\bf 76.3} & \cellcolor{highlight}{\bf 72.4} & \cellcolor{highlight}{\bf 49.7} & \cellcolor{highlight}{\bf 38.1} & \cellcolor{highlight}{\bf 8.8} & \cellcolor{highlight}{\bf 41.6} & \cellcolor{highlight}\textbf{47.8} \\

\midrule
\multicolumn{8}{c}{LLaMA2-13B-3bit-g64} \\
\midrule
FP16 & 82.1 & 76.6 & 59.4 & 37.4 & 22.5 & 55.7 & 55.6 \\
QAT & 80.7 & 75.1 & 55.5 & {\bf 39.0} & 16.8 & 52.9 & 53.3 \\
GPTQ & 79.2 & 74.4 & 56.5 & 36.0 & 16.4 & 52.4 & 52.5 \\
\cellcolor{highlight}\textbf{CherryQ} & \cellcolor{highlight}{\bf 81.1} & \cellcolor{highlight}{\bf 76.2} & \cellcolor{highlight}{\bf 57.3} & \cellcolor{highlight}38.0 & \cellcolor{highlight}{\bf 18.4} & \cellcolor{highlight}{\bf 53.5} & \cellcolor{highlight}\textbf{54.1} \\

\midrule
\multicolumn{8}{c}{LLaMA2-13B-3bit-g128} \\
\midrule
FP16 & 82.1 & 76.6 & 59.4 & 37.4 & 22.5 & 55.7 & 55.6 \\
QAT & 80.7 & {\bf 75.5} & 55.3 & 38.8 & 16.0 & 51.9 & 53.0 \\
GPTQ & 79.1 & 75.4 & 54.1 & 34.9 & 15.6 & 50.3 & 51.6 \\
\cellcolor{highlight}\textbf{CherryQ} & \cellcolor{highlight}{\bf 81.0} & \cellcolor{highlight}75.4 & \cellcolor{highlight}{\bf 56.7} & \cellcolor{highlight}{\bf 38.9} & \cellcolor{highlight}{\bf 17.8} & \cellcolor{highlight}{\bf 52.5} & \cellcolor{highlight}\textbf{53.7} \\

\bottomrule
\end{tabular}
\end{table}

\begin{table}[htbp]
\centering
\caption{Performance comparison of different 4-bit quantization methods for LLaMA2-7B and LLaMA2-13B models over Huggingface OpenLLM Leaderboard.}
\label{tab:quantization_comparison_4bit_adjusted}
\begin{tabular}{l|cccccc|c}
\toprule
Method & Hellaswag & Winogrande & ARC & TruthfulQA & GSM8K & MMLU & Average (\textcolor{red}{$\uparrow$}) \\
\midrule
\multicolumn{8}{c}{LLaMA2-7B-4bit-g128} \\
\midrule
FP16 & 78.6 & 74.0 & 53.2 & 38.8 & 14.5 & 46.7 & 51.0 \\
QAT & 77.5 & 72.2 & {\bf 52.0} & 39.0 & 10.6 & 43.7 & 49.2 \\
GPTQ & 77.6 & 72.9 & {\bf 52.0} & 39.1 & 11.1 & 43.8 & 49.4 \\
\cellcolor{highlight}\textbf{CherryQ} & \cellcolor{highlight}{\bf 77.8} & \cellcolor{highlight}{\bf 73.5} & \cellcolor{highlight}51.5 & \cellcolor{highlight}{\bf 39.5} & \cellcolor{highlight}{\bf 12.9} & \cellcolor{highlight}{\bf 44.4} & \cellcolor{highlight}\textbf{49.9} \\
\midrule
\multicolumn{8}{c}{LLaMA2-13B-4bit-g128} \\
\midrule
FP16 & 82.1 & 76.6 & 59.4 & 37.4 & 22.5 & 55.7 & 55.6 \\
QAT & 81.9 & 75.7 & 57.9 & {\bf 38.9} & 19.6 & 54.2 & 54.7 \\
GPTQ & 81.5 & 76.8 & 57.4 & 36.1 & 20.4 & 54.6 & 54.5 \\
\cellcolor{highlight}\textbf{CherryQ} & \cellcolor{highlight}{\bf 82.0} & \cellcolor{highlight}{\bf 77.0} & \cellcolor{highlight}{\bf 58.6} & \cellcolor{highlight}38.8 & \cellcolor{highlight}{\bf 21.0} & \cellcolor{highlight}{\bf 54.6} & \cellcolor{highlight}\textbf{55.3} \\
\bottomrule
\end{tabular}
\end{table}

\subsection{Effect of Chat LLM Quantization}

%In this subsection, we investigate the effect of quantization on open-ended chat models. We focus on evaluating the effectiveness of different quantization methods in preserving the quality of generated responses compared to the full-precision (FP16) models.

We conducted experiments on Vicuna-1.5~\cite{chiang2023vicuna}. We apply 3-bit quantization with group size=128 for CherryQ and other baselines.

{\bf Evaluation} To assess the performance of quantized open-ended chat models, we employ a pairwise comparison on the Vicuna-bench~\cite{zheng2024judging}, which consists of 80 test samples. We compare the responses generated by the quantized models against those generated by the original 16-bit Vicuna-1.5. The evaluation is performed using GPT-4, which automatically classifies the quantized model's response as ``win'', ``tie'', or ``lose'' relative to the FP16 model's response. To get rid of the ordering effect of the evaluation, we follow~\cite{lin2023awq} to compare the responses with both orders, resulting in 160 trials.

\begin{figure*}[t]
\centering
\includegraphics[width=1.0\textwidth]{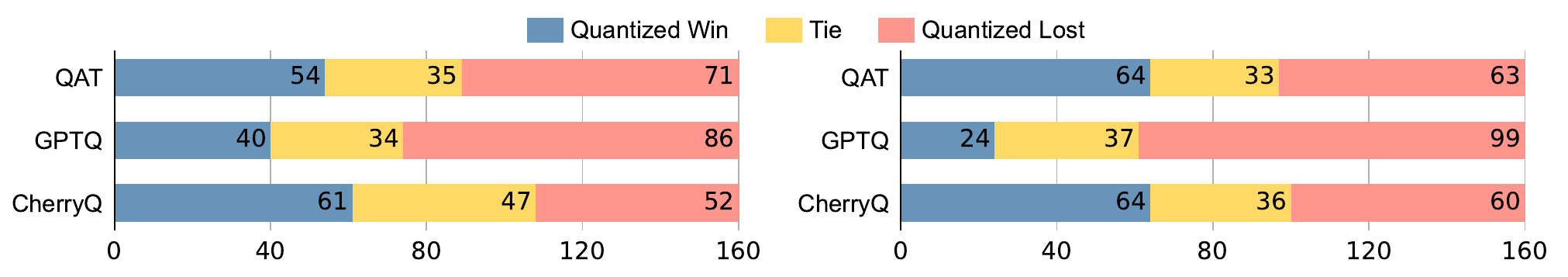}
\caption{Comparison of 3-bit quantized models to FP16 Vicuna-1.5. (Left) Comparisons to Vicuna-1.5-7B. (Right) Comparisons to Vicuna-1.5-13B. CherryQ even shows competitive quality compared to the 16-bit counterpart.}
\label{fig:vicuna_bench}
\end{figure*}

Figure~\ref{fig:vicuna_bench} presents the results of the pairwise comparison for each quantized model against its FP16 counterpart. The results demonstrate that CherryQ consistently outperforms other quantization baselines in preserving the performance of chat models. It achieves the highest number of wins and ties against the FP16 models, while minimizing the number of losses.

Notably, {\bf 3-bit CherryQ achieves a slightly better win-tie-lose ratio over the FP16 Vicuna model}, indicating that the 3-bit quantized model performs on par with or even better than the FP16 model. As intuitively CherryQ cannot surpass the target 16 bit model, we think the result suggests that CherryQ maintains almost all its performance even at 3 bit, making GPT-4 hard to distinguish the quality of low-bit and FP16 models.

%The superior performance of CherryQ can be attributed to its ability to identify and preserve the critical "cherry" parameters during quantization. By minimizing the quantization error for these influential parameters, CherryQ ensures that the quantized models retain their expressiveness and generate high-quality responses.

%These findings highlight the effectiveness of CherryQ in quantizing chat models while maintaining their performance, paving the way for more efficient deployment and utilization of open-ended dialogue systems.

\subsection{Extreme 2-Bit Quantization}

We further explore the extreme case of 2-bit quantization. Although 2-bit quantization greatly reduces memory requirements for model storage and inference, existing methods still show a significant performance gap compared to their 16-bit counterparts.

{\bf Implementation Details} To achieve high-quality 2-bit quantization, we integrated the scaling-up trick introduced in \cite{lin2023awq}. Specifically, after identifying cherry and normal parameters, we automatically search for the optimal scale of each column of normal parameters that minimizes the output difference after quantization for each layer. The quantization function is formulated as $Q'(w) = Q(w \cdot s)/s$, where $Q(\cdot)$ represents standard asymmetric quantization on the min-max grid, and $s$ is a constant that scales up the normal parameters and remains fixed during the training process. The cherry parameters are excluded from quantization and retain their 16-bit precision throughout the grid search.

\begin{table}[h]
\centering
\small
\setlength{\tabcolsep}{4pt}
\caption{Perplexity (\textcolor{red}{$\downarrow$}) of 2-bit quantization on LLaMA2 models. The
results of GPTQ, AWQ and OmniQuant are from~\cite{shao2023omniquant}.}
\label{tab:ablation_model_comparison_2bit}
\begin{tabular}{lcccccccccccc}
\toprule
Method & \multirow{2}{*}{\shortstack{Avg.\\bit}} & \multicolumn{2}{c}{7B-2bit-g128} & \multirow{2}{*}{\shortstack{Avg.\\bit}} & \multicolumn{2}{c}{7B-2bit-g64} & \multirow{2}{*}{\shortstack{Avg.\\bit}} & \multicolumn{2}{c}{13B-2bit-g128} & \multirow{2}{*}{\shortstack{Avg.\\bit}} & \multicolumn{2}{c}{13B-2bit-g64} \\
\cmidrule(lr){3-4} \cmidrule(lr){6-7} \cmidrule(lr){9-10} \cmidrule(lr){12-13}
& & c4 & wiki2 & & c4 & wiki2 & & c4 & wiki2 & & c4 & wiki2 \\
\midrule
FP16 & 16 & 6.97 & 5.47 & 16 & 6.97 & 5.47 & 16 & 6.47 & 4.88 & 16 & 6.47 & 4.88 \\
GPTQ & 2.15 & 33.70 & 36.77 & 2.30 & 19.40 & 20.85 & 2.15 & 20.97 & 28.14 & 2.30 & 12.48 & 22.44 \\
AWQ & 2.15 & $> 10^5$ & $>10^5$ & 2.30 & $>10^5$ & $>10^5$ & 2.15 & $>10^4$ & $>10^5$ & 2.30 & $>10^4$ & $>10^5$ \\
OmniQuant & 2.15 & 15.02 & 11.06 & 2.30 & 12.72 & 9.62 & 2.15 & 11.05 & 8.26 & 2.30 & 10.05 & 7.56 \\
\cellcolor{highlight}\textbf{CherryQ} & 
\cellcolor{highlight}2.19 & 
\cellcolor{highlight}\textbf{9.55} & \cellcolor{highlight}\textbf{8.34} & 
\cellcolor{highlight}2.34 & 
\cellcolor{highlight}\textbf{9.08} & \cellcolor{highlight}\textbf{7.84} & 
\cellcolor{highlight}2.19 & 
\cellcolor{highlight}\textbf{8.40} & \cellcolor{highlight}\textbf{7.20} & 
\cellcolor{highlight}2.33 & 
\cellcolor{highlight}\textbf{8.02} & \cellcolor{highlight}\textbf{6.72} \\
\bottomrule
\end{tabular}
\label{tab:exp:2bit}
\end{table}

{\bf Results} Table~\ref{tab:exp:2bit} presents the perplexities of 2-bit quantization on LLAMA2 models. Compared to existing methods such as GPTQ, AWQ, and OmniQuant, our proposed CherryQ method demonstrates superior performance across all metrics. Specifically, CherryQ achieves perplexity scores of 9.55 and 8.34 in the 7B-3bit-g128 and 7B-3bit-g64 settings, respectively. These results significantly outperform other methods, validating the effectiveness of CherryQ in 2-bit quantization.

% \textcolor{red}{Within the framework of mixed-precision training, CherryQ is compatible with essentially other quantization functions, e.g. asymmetric quantization, as long as the normal parameters can be correctly updated through the Straight-Through Estimator (STE). CherryQ is also orthogonal to other Post-Training Quantization (PTQ) methods, as PTQ essentially provides CherryQ with an optimized set of initial training parameters. From this perspective, we can push the limit of CherryQ by integrating the scaling-up trick introduced in \cite{lin2023awq}, which is both effective and simple to implement.}

% \textcolor{red}{Specifically, after identifying cherry and normal parameters, we automatically search for the optimal scale of each column of normal parameters that minimizes the output difference after quantization for a specific layer as detailed in \cite{lin2023awq}. The cherry parameters are excluded from quantization and maintains their 16-bit precision during the grid search. The quantization function is formulated as:}

% $Q'(w) = Q(w\cdot s)/s$

% \textcolor{red}{where $Q(\cdot)$ is the standard asymmetric quantization on the min-max grid since the scales obtained through the grid search are predominantly close to 1 if symmetric quantization is applied, indicating the absence of any scaling-up effect. $s$ is the constant that scales up the normal parameters and is frozen during the training process.}

\subsection{Comparison of Parameter Selection Criteria}
\label{sec:exp:criteria}

To evaluate the effectiveness of our proposed impact-based parameter selection criterion, we conducted experiments comparing it with the criterions of parameter weights and activations. Table~\ref{tab:ablation_model_comparison_3bit} presents the perplexity of LLaMA2-7B-3bit and LLaMA2-13B-3bit models, using both criteria for cherry parameter selection.

From the results, it is evident that the impact-based criterion consistently outperforms other criterions across all settings. %For LLaMA2-7B-3bit, using impact with a group size of 64 achieves perplexities of 7.34 and 5.87 on C4 and WikiText-2, respectively, which are significantly lower than the corresponding perplexities of 7.93 and 6.40 obtained using magnitude. Similar results can be found for other settings (LLaMA2-13B-3bit and group size $=64$).
These results demonstrate that our proposed impact-based criterion is a more effective measure to identify cherry parameters. The impacts identify and preserve the most critical parameters during the quantization process. These results are consistent with the analysis in Sec~\ref{sec:phenomenon} regarding the effectiveness of heterogeneity scores in selecting cherry parameters.
%We think this justify the heterogeneity of parameter impacts against parameter magnitudes as we highlighted in \S~\ref{sec:phenomenon}.

\begin{table}[h]
\centering
\caption{Perplexity (\textcolor{red}{$\downarrow$}) of different parameter selection criteria.}
\label{tab:ablation_model_comparison_3bit}
\begin{tabular}{lcccc}
\toprule
 Method & \multicolumn{2}{c}{LLaMA2-7B-3bit} & \multicolumn{2}{c}{LLaMA2-13B-3bit} \\
\cmidrule(lr){2-3} \cmidrule(lr){4-5} & c4 & wiki2 & c4 & wiki2 \\
\midrule
Weight-g64 & 7.93 & 6.40 & 6.91 & 5.35 \\
Activation-g64 & 7.37 & 5.89 & 6.77 & 5.22 \\
\cellcolor{highlight}{\bf Impact-g64} & \cellcolor{highlight}{\bf 7.34} & \cellcolor{highlight}{\bf 5.87} & \cellcolor{highlight}{\bf 6.76} & \cellcolor{highlight}{\bf 5.21} \\
\midrule
Weight-g128 & 8.12 & 6.58 & 6.94 & 5.37 \\
Activation-g128 & 7.51 & 6.03 & 6.81 & 5.27 \\
\cellcolor{highlight}{\bf Impact-g128} & \cellcolor{highlight}{\bf 7.39} & \cellcolor{highlight}{\bf 5.93} & \cellcolor{highlight}{\bf 6.80} & \cellcolor{highlight}{\bf 5.26} \\
\midrule
 & \multicolumn{2}{c}{LLaMA2-7B-4bit} & \multicolumn{2}{c}{LLaMA2-13B-4bit} \\
\midrule
Weight-g128 & 7.19 & 5.68 & 6.62 & 5.05 \\
Activation-g128 & 7.09 & 5.59 & 6.56 & 5.00 \\
\cellcolor{highlight}{\bf Impact-g128} & \cellcolor{highlight}{\bf 7.07} & \cellcolor{highlight}{\bf 5.58} & \cellcolor{highlight}{\bf 6.56} & \cellcolor{highlight}{\bf 4.99} \\
\bottomrule
\end{tabular}
\end{table}

%% file: conclu.tex
\section{Conclusion}

%In this paper, we investigated the parameter heterogeneity phenomenon in LLMs. Our experiments on LLaMA2, Mistral, Gemma, and Vicuna models, consistently demonstrated that a small subset of parameters plays a crucial role in maintaining the model's performance, while the majority of parameters can be quantized to ultra-low precision without significant degradation. This finding highlights the potential for efficient model compression and quantization techniques that take into account the heterogeneous nature of parameter importance.

%Motivated by this observation, we proposed a novel impact-based parameter selection criterion for quantization. Our method effectively identifies and preserves the most critical cherry parameters during the quantization process. We use a QAT framework for unified optimization of both cherry parameters and normal parameters. Extensive experiments demonstrate that CherryQ outperforms the commonly used magnitude-based criterion, achieving significantly lower perplexity scores and better downstream performance. The heterogeneity and proposed approach pave the way for more efficient deployment of LLMs in resource-constrained environments.

In this paper, we systematically investigated the phenomenon of parameter heterogeneity in large language models (LLMs). Our experiments on LLaMA2, Mistral, Gemma, and Vicuna models consistently demonstrated that a small subset of parameters, referred to as "cherry" parameters, play a crucial role in maintaining the model's performance, while the vast majority of parameters can be quantized to ultra-low precision without significant degradation. This finding highlights the potential of the heterogeneous nature of parameter importance.

Motivated by this observation, we proposed a novel %impact-based parameter selection criterion for quantization. Our method effectively identifies and preserves the most critical cherry parameters during the quantization process. We introduced the 
CherryQ quantization algorithm, which uses a quantization-aware training framework for the end-to-end optimization of both cherry parameters and normal parameters. Extensive experiments demonstrate that CherryQ achieves significantly lower perplexity scores and better downstream performance.

\section{Limitations}
\label{sec:limitations}

There are some limitations to consider. First, the method relies heavily on the accurate identification of cherry parameters, which may vary across different model architectures and training datasets. This dependency could potentially limit the generalization ability of CherryQ to new or unseen models. Second, the computational overhead required for the impact-based identification and scaling of parameters, although justified by the performance gains, may pose challenges for extremely large models or those deployed in real-time systems with stringent latency requirements.% Lastly, our experiments are primarily conducted on specific benchmark datasets and models, and further validation across a broader range of applications and environments is necessary to fully ascertain the robustness and scalability of CherryQ .

%% file: appendix.tex
\appendix

\section{Effect of Chat LLM Quantization on MMLU}
We further evaluate the performance of CherryQ on the MMLU benchmark by quantizing the Vicuna1.5 model. As shown in Table~\ref{tab:mmlu}, CherryQ outperforms both QAT and GPTQ in terms of average accuracy across almost all categories.

\begin{table}[h]
\centering
\caption{Comparison of different 3-bit quantization methods on zero-shot MMLU accuracy applied to Vicuna-1.5.}
\label{tab:mmlu}
\begin{tabular}{l|cccc|c}
\toprule
Method & Humanities & STEM & Social Sciences & Other & Average \\
\midrule
\multicolumn{6}{c}{Vicuna1.5-7B-3bit-g128} \\
\midrule
FP16 & 46.8 & 39.4 & 57.9 & 56.3 & 49.9 \\
QAT & 43.4 & {\bf 37.7} & 53.0 & 52.4 & 46.4 \\
GPTQ & 42.7 & 37.3 & 53.0 & 51.0 & 45.7 \\
\cellcolor{highlight}\textbf{CherryQ} & \cellcolor{highlight}{\bf 43.8} & \cellcolor{highlight}37.2 & \cellcolor{highlight}{\bf 54.3} & \cellcolor{highlight}{\bf 53.5} & \cellcolor{highlight}\textbf{46.9} \\
\midrule
\multicolumn{6}{c}{Vicuna1.5-13B-3bit-g128} \\
\midrule
FP16 & 50.2 & 43.5 & 63.0 & 62.0 & 54.3 \\
QAT & 47.8 & 40.1 & 58.6 & 58.1 & 50.9 \\
GPTQ & 46.1 & 39.4 & 57.6 & 55.2 & 49.3 \\
\cellcolor{highlight}\textbf{CherryQ} & \cellcolor{highlight}{\bf 49.0} & \cellcolor{highlight}{\bf 40.6} & \cellcolor{highlight}{\bf 60.2} & \cellcolor{highlight}{\bf 58.8} & \cellcolor{highlight}\textbf{51.9} \\
\bottomrule
\end{tabular}
\end{table}

\section{Training Details}
\label{sec:append:details}
For all LLM scales (7B, 13B), and both base models and chat models (LLaMA2, Vicuna-v1.5), we train the models on a single node with 8 x A100 80GiB GPUs. We use a total batch size of 128, a learning rate of 2e-5, a weight decay of 0.0, a cosine scheduler with 5\% warm-up steps. The final learning rate is 25\% of the peak learning rate for 2/3-bit LLMs, 10\% for 4-bit LLMs. We train 1 epoch on base models, 2 epochs on chat models.

\section{Parameter Representation Details}
\label{sec:append:quant}
For normal parameters, we employ {\it full range symmetric MinMax quantization} to quantize their weights~\cite{krishnamoorthi2018quantizing}. Specifically, an FP16 value is mapped to the range of $[-2^{k-1},2^{k-1}-1]$ and symmetrically distributed on both sides of the coordinate axis. The quantization of an FP16 tensor $X^{FP16}$ to $k$ bits is computed by:
\begin{equation}
X^{Intk}=\lfloor Clip(\frac{X^{FP16}}{S}, -2^{k-1}+\epsilon, 2^{k-1}-\epsilon)-0.5 \rceil  
\end{equation}
where $\lfloor \cdot \rceil$ denotes the round function, $S$ is the quantization scaling factor $S = \frac{Max(|X^{FP16}|)}{2^{k-1}}$, and $\epsilon$ is a very small positive number (= 0.01 in our setting) to ensure that $X^{Intk}$ falls into the target range.

Dequantization restores the quantized integer values based on the scaling factor:

\begin{equation}
Dequant(S, X^{Intk})=S(X^{Intk}+0.5)
\end{equation}

To further improve the quantization accuracy, we adopt a widely-used parameter grouping strategy. Specifically, the parameters are divided into groups in order, and each group independently calculates its scaling factor. For example, if we divide a parameter matrix $W \in \mathbb{R}^{r\times c}$ that needs to be quantized with a group size of $B$, we will obtain a total of $r \times (c / B)$ groups.

\section{Licenses for Existing Assets}
\label{sec:append:license}
We list the assets used in this paper and their licenses below:
\begin{itemize}
    \item
    \cite{chiang2023vicuna}, llama2
    \item https://huggingface.co/TheBloke, llama2
    \item https://github.com/AutoGPTQ/AutoGPTQ, MIT License
    \item
    \cite{touvron2023llama}, arXiv.org perpetual, non-exclusive license 1.0
    \item
    \cite{liu2023llm}, arXiv.org perpetual, non-exclusive license 1.0
    \item
    \cite{raffel2020exploring}, arXiv.org perpetual, non-exclusive license 1.0
    \item
    \cite{merity2016pointer}, arXiv.org perpetual, non-exclusive license 1.0
\end{itemize}

%% file: checklist.tex
\newpage
\section*{NeurIPS Paper Checklist}

\begin{enumerate}

\item {\bf Claims}
    \item[] Question: Do the main claims made in the abstract and introduction accurately reflect the paper's contributions and scope?
    \item[] Answer: \answerYes{} % Replace by \answerYes{}, \answerNo{}, or \answerNA{}.
    \item[] Justification: We have fully explained the motivation and contributions of the paper in the introduction section, and have verified them in the experimental section.

\item {\bf Limitations}
    \item[] Question: Does the paper discuss the limitations of the work performed by the authors?
    \item[] Answer: \answerYes{} % Replace by \answerYes{}, \answerNo{}, or \answerNA{}.
    \item[] Justification: Please refer to Section~\ref{sec:limitations}.

\item {\bf Theory Assumptions and Proofs}
    \item[] Question: For each theoretical result, does the paper provide the full set of assumptions and a complete (and correct) proof?
    \item[] Answer: \answerYes{} % Replace by \answerYes{}, \answerNo{}, or \answerNA{}.
    \item[] Justification: Please refer to Section~\ref{sec:measure}.

    \item {\bf Experimental Result Reproducibility}
    \item[] Question: Does the paper fully disclose all the information needed to reproduce the main experimental results of the paper to the extent that it affects the main claims and/or conclusions of the paper (regardless of whether the code and data are provided or not)?
    \item[] Answer: \answerYes{} % Replace by \answerYes{}, \answerNo{}, or \answerNA{}.
    \item[] Justification: Please refer to Section~\ref{sec:exp:implement_details}, Section~\ref{sec:append:details}, and the attached code.

\item {\bf Open access to data and code}
    \item[] Question: Does the paper provide open access to the data and code, with sufficient instructions to faithfully reproduce the main experimental results, as described in supplemental material?
    \item[] Answer: \answerYes{} % Replace by \answerYes{}, \answerNo{}, or \answerNA{}.
    \item[] Justification: We have attached the codes in the submission.

\item {\bf Experimental Setting/Details}
    \item[] Question: Does the paper specify all the training and test details (e.g., data splits, hyperparameters, how they were chosen, type of optimizer, etc.) necessary to understand the results?
    \item[] Answer: \answerYes{} % Replace by \answerYes{}, \answerNo{}, or \answerNA{}.
    \item[] Justification: Please refer to Section~\ref{sec:exp:implement_details}, Section~\ref{sec:append:details}, and the attached code.

\item {\bf Experiment Statistical Significance}
    \item[] Question: Does the paper report error bars suitably and correctly defined or other appropriate information about the statistical significance of the experiments?
    \item[] Answer: \answerNo{}{} % Replace by \answerYes{}, \answerNo{}, or \answerNA{}.
    \item[] Justification: Error bars are not reported because it would be too computationally expensive.

\item {\bf Experiments Compute Resources}
    \item[] Question: For each experiment, does the paper provide sufficient information on the computer resources (type of compute workers, memory, time of execution) needed to reproduce the experiments?
    \item[] Answer: \answerYes{} % Replace by \answerYes{}, \answerNo{}, or \answerNA{}.
    \item[] Justification: All experiments were completed on nodes with the GPUs of 8*A100 80G. Please refer to Section~\ref{sec:exp:implement_details}.
    
\item {\bf Code Of Ethics}
    \item[] Question: Does the research conducted in the paper conform, in every respect, with the NeurIPS Code of Ethics \url{https://neurips.cc/public/EthicsGuidelines}?
    \item[] Answer: \answerYes{} % Replace by \answerYes{}, \answerNo{}, or \answerNA{}.
    \item[] Justification: We carefully read and follow the NeurIPS Code of Ethics.

\item {\bf Broader Impacts}
    \item[] Question: Does the paper discuss both potential positive societal impacts and negative societal impacts of the work performed?
    \item[] Answer: \answerNo{} % Replace by \answerYes{}, \answerNo{}, or \answerNA{}.
    \item[] Justification: The paper is purely fundamental research and does not involve social impact.
    
\item {\bf Safeguards}
    \item[] Question: Does the paper describe safeguards that have been put in place for responsible release of data or models that have a high risk for misuse (e.g., pretrained language models, image generators, or scraped datasets)?
    \item[] Answer: \answerNo{}{} % Replace by \answerYes{}, \answerNo{}, or \answerNA{}.
    \item[] Justification: This paper does not release new models or datasets.

\item {\bf Licenses for existing assets}
    \item[] Question: Are the creators or original owners of assets (e.g., code, data, models), used in the paper, properly credited and are the license and terms of use explicitly mentioned and properly respected?
    \item[] Answer: \answerYes{} % Replace by \answerYes{}, \answerNo{}, or \answerNA{}.
    \item[] Justification: Please refer to Section~\ref{sec:append:license}

\item {\bf New Assets}
    \item[] Question: Are new assets introduced in the paper well documented and is the documentation provided alongside the assets?
    \item[] Answer: \answerYes{}{} % Replace by \answerYes{}, \answerNo{}, or \answerNA{}.
    \item[] Justification: We have attached codes with documents.

\item {\bf Crowdsourcing and Research with Human Subjects}
    \item[] Question: For crowdsourcing experiments and research with human subjects, does the paper include the full text of instructions given to participants and screenshots, if applicable, as well as details about compensation (if any)? 
    \item[] Answer: \answerNA{} % Replace by \answerYes{}, \answerNo{}, or \answerNA{}.
    \item[] Justification: The paper does not involve crowdsourcing nor research with human subjects.

\item {\bf Institutional Review Board (IRB) Approvals or Equivalent for Research with Human Subjects}
    \item[] Question: Does the paper describe potential risks incurred by study participants, whether such risks were disclosed to the subjects, and whether Institutional Review Board (IRB) approvals (or an equivalent approval/review based on the requirements of your country or institution) were obtained?
    \item[] Answer: \answerNA{}{} % Replace by \answerYes{}, \answerNo{}, or \answerNA{}.
    \item[] Justification: The paper does not involve crowdsourcing nor research with human subjects.

\end{enumerate}

%% file: main.bbl
\begin{thebibliography}{10}

\bibitem{achiam2023gpt}
Josh Achiam, Steven Adler, Sandhini Agarwal, Lama Ahmad, Ilge Akkaya, Florencia~Leoni Aleman, Diogo Almeida, Janko Altenschmidt, Sam Altman, Shyamal Anadkat, et~al.
\newblock Gpt-4 technical report.
\newblock {\em arXiv preprint arXiv:2303.08774}, 2023.

\bibitem{bai2023qwen}
Jinze Bai, Shuai Bai, Yunfei Chu, Zeyu Cui, Kai Dang, Xiaodong Deng, Yang Fan, Wenbin Ge, Yu~Han, Fei Huang, et~al.
\newblock Qwen technical report.
\newblock {\em arXiv preprint arXiv:2309.16609}, 2023.

\bibitem{bengio2013estimating}
Yoshua Bengio, Nicholas L{\'e}onard, and Aaron Courville.
\newblock Estimating or propagating gradients through stochastic neurons for conditional computation.
\newblock {\em arXiv preprint arXiv:1308.3432}, 2013.

\bibitem{bondarenko2021understanding}
Yelysei Bondarenko, Markus Nagel, and Tijmen Blankevoort.
\newblock Understanding and overcoming the challenges of efficient transformer quantization.
\newblock In {\em Proceedings of the 2021 Conference on Empirical Methods in Natural Language Processing}, pages 7947--7969, 2021.

\bibitem{chiang2023vicuna}
Wei-Lin Chiang, Zhuohan Li, Zi~Lin, Ying Sheng, Zhanghao Wu, Hao Zhang, Lianmin Zheng, Siyuan Zhuang, Yonghao Zhuang, Joseph~E Gonzalez, et~al.
\newblock Vicuna: An open-source chatbot impressing gpt-4 with 90\%* chatgpt quality, march 2023.
\newblock {\em URL https://lmsys. org/blog/2023-03-30-vicuna}, 3(5), 2023.

\bibitem{dettmers2022gpt3}
Tim Dettmers, Mike Lewis, Younes Belkada, and Luke Zettlemoyer.
\newblock Gpt3. int8 (): 8-bit matrix multiplication for transformers at scale.
\newblock {\em Advances in Neural Information Processing Systems}, 35:30318--30332, 2022.

\bibitem{dettmers2024qlora}
Tim Dettmers, Artidoro Pagnoni, Ari Holtzman, and Luke Zettlemoyer.
\newblock Qlora: Efficient finetuning of quantized llms.
\newblock {\em Advances in Neural Information Processing Systems}, 36, 2024.

\bibitem{dettmers2023spqr}
Tim Dettmers, Ruslan~A Svirschevski, Vage Egiazarian, Denis Kuznedelev, Elias Frantar, Saleh Ashkboos, Alexander Borzunov, Torsten Hoefler, and Dan Alistarh.
\newblock Spqr: A sparse-quantized representation for near-lossless llm weight compression.
\newblock In {\em The Twelfth International Conference on Learning Representations}, 2023.

\bibitem{frantar2023gptq}
Elias Frantar, Saleh Ashkboos, Torsten Hoefler, and Dan Alistarh.
\newblock Gptq: Accurate post-training quantization for generative pre-trained transformers.
\newblock In {\em The Eleventh International Conference on Learning Representations}, 2023.

\bibitem{hassibi1993optimal}
Babak Hassibi, David~G Stork, and Gregory~J Wolff.
\newblock Optimal brain surgeon and general network pruning.
\newblock In {\em IEEE international conference on neural networks}, pages 293--299. IEEE, 1993.

\bibitem{jiang2023mistral}
Albert~Q Jiang, Alexandre Sablayrolles, Arthur Mensch, Chris Bamford, Devendra~Singh Chaplot, Diego de~las Casas, Florian Bressand, Gianna Lengyel, Guillaume Lample, Lucile Saulnier, et~al.
\newblock Mistral 7b.
\newblock {\em arXiv preprint arXiv:2310.06825}, 2023.

\bibitem{kim2023squeezellm}
Sehoon Kim, Coleman Hooper, Amir Gholami, Zhen Dong, Xiuyu Li, Sheng Shen, Michael~W Mahoney, and Kurt Keutzer.
\newblock Squeezellm: Dense-and-sparse quantization.
\newblock {\em arXiv preprint arXiv:2306.07629}, 2023.

\bibitem{kovaleva2021bert}
Olga Kovaleva, Saurabh Kulshreshtha, Anna Rogers, and Anna Rumshisky.
\newblock Bert busters: Outlier dimensions that disrupt transformers.
\newblock {\em Findings of the Association for Computational Linguistics: ACL-IJCNLP 2021}, 2021.

\bibitem{krishnamoorthi2018quantizing}
Raghuraman Krishnamoorthi.
\newblock Quantizing deep convolutional networks for efficient inference: A whitepaper.
\newblock {\em arXiv preprint arXiv:1806.08342}, 2018.

\bibitem{lecun1989optimal}
Yann LeCun, John Denker, and Sara Solla.
\newblock Optimal brain damage.
\newblock {\em Advances in neural information processing systems}, 2, 1989.

\bibitem{li2020brecq}
Yuhang Li, Ruihao Gong, Xu~Tan, Yang Yang, Peng Hu, Qi~Zhang, Fengwei Yu, Wei Wang, and Shi Gu.
\newblock Brecq: Pushing the limit of post-training quantization by block reconstruction.
\newblock In {\em International Conference on Learning Representations}, 2020.

\bibitem{lin2023awq}
Ji~Lin, Jiaming Tang, Haotian Tang, Shang Yang, Xingyu Dang, and Song Han.
\newblock Awq: Activation-aware weight quantization for llm compression and acceleration.
\newblock {\em arXiv preprint arXiv:2306.00978}, 2023.

\bibitem{liu2023llm}
Zechun Liu, Barlas Oguz, Changsheng Zhao, Ernie Chang, Pierre Stock, Yashar Mehdad, Yangyang Shi, Raghuraman Krishnamoorthi, and Vikas Chandra.
\newblock Llm-qat: Data-free quantization aware training for large language models.
\newblock {\em arXiv preprint arXiv:2305.17888}, 2023.

\bibitem{merity2016pointer}
Stephen Merity, Caiming Xiong, James Bradbury, and Richard Socher.
\newblock Pointer sentinel mixture models.
\newblock {\em arXiv preprint arXiv:1609.07843}, 2016.

\bibitem{raffel2020exploring}
Colin Raffel, Noam Shazeer, Adam Roberts, Katherine Lee, Sharan Narang, Michael Matena, Yanqi Zhou, Wei Li, and Peter~J Liu.
\newblock Exploring the limits of transfer learning with a unified text-to-text transformer.
\newblock {\em Journal of machine learning research}, 21(140):1--67, 2020.

\bibitem{shao2023omniquant}
Wenqi Shao, Mengzhao Chen, Zhaoyang Zhang, Peng Xu, Lirui Zhao, Zhiqian Li, Kaipeng Zhang, Peng Gao, Yu~Qiao, and Ping Luo.
\newblock Omniquant: Omnidirectionally calibrated quantization for large language models.
\newblock In {\em The Twelfth International Conference on Learning Representations}, 2023.

\bibitem{team2024gemma}
Gemma Team, Thomas Mesnard, Cassidy Hardin, Robert Dadashi, Surya Bhupatiraju, Shreya Pathak, Laurent Sifre, Morgane Rivi{\`e}re, Mihir~Sanjay Kale, Juliette Love, et~al.
\newblock Gemma: Open models based on gemini research and technology.
\newblock {\em arXiv preprint arXiv:2403.08295}, 2024.

\bibitem{touvron2023llama}
Hugo Touvron, Louis Martin, Kevin Stone, Peter Albert, Amjad Almahairi, Yasmine Babaei, Nikolay Bashlykov, Soumya Batra, Prajjwal Bhargava, Shruti Bhosale, et~al.
\newblock Llama 2: Open foundation and fine-tuned chat models.
\newblock {\em arXiv preprint arXiv:2307.09288}, 2023.

\bibitem{wei2023outlier}
Xiuying Wei, Yunchen Zhang, Yuhang Li, Xiangguo Zhang, Ruihao Gong, Jinyang Guo, and Xianglong Liu.
\newblock Outlier suppression+: Accurate quantization of large language models by equivalent and effective shifting and scaling.
\newblock In {\em Proceedings of the 2023 Conference on Empirical Methods in Natural Language Processing}, pages 1648--1665, 2023.

\bibitem{wei2022outlier}
Xiuying Wei, Yunchen Zhang, Xiangguo Zhang, Ruihao Gong, Shanghang Zhang, Qi~Zhang, Fengwei Yu, and Xianglong Liu.
\newblock Outlier suppression: Pushing the limit of low-bit transformer language models.
\newblock {\em Advances in Neural Information Processing Systems}, 35:17402--17414, 2022.

\bibitem{xiao2023smoothquant}
Guangxuan Xiao, Ji~Lin, Mickael Seznec, Hao Wu, Julien Demouth, and Song Han.
\newblock Smoothquant: Accurate and efficient post-training quantization for large language models.
\newblock In {\em International Conference on Machine Learning}, pages 38087--38099. PMLR, 2023.

\bibitem{zheng2024judging}
Lianmin Zheng, Wei-Lin Chiang, Ying Sheng, Siyuan Zhuang, Zhanghao Wu, Yonghao Zhuang, Zi~Lin, Zhuohan Li, Dacheng Li, Eric Xing, et~al.
\newblock Judging llm-as-a-judge with mt-bench and chatbot arena.
\newblock {\em Advances in Neural Information Processing Systems}, 36, 2024.

\end{thebibliography}
